\documentclass[preprint,1p,times,onecolumn]{elsarticle}

\usepackage[outdir=./]{epstopdf}%
\usepackage[caption=false]{subfig}%
\usepackage{amssymb}
\usepackage{amsmath,graphicx}
\usepackage{graphicx}
\usepackage{ragged2e}
\usepackage{multirow}
\usepackage{flushend}
\usepackage[table,xcdraw]{xcolor}
\usepackage{comment}
\usepackage{xcolor}
\definecolor{darkgray}{rgb}{0.3, 0.3, 0.3}
\usepackage{xspace}
\usepackage{svg}
\usepackage{algorithm}
\usepackage{algpseudocode}
\usepackage{amsmath}
\usepackage{pgf-pie}
\usepackage{pgfplots}
\usepackage{adjustbox}

\usepackage{pgfplotstable}

\pgfplotstableset{
    color cells/.style={
        col sep=comma,
        string type,
        postproc cell content/.code={%
                \pgfkeysalso{@cell content=\rule{0cm}{2.4ex}\cellcolor{black!##1}\pgfmathtruncatemacro\number{##1}\ifnum\number>50\color{white}\fi##1}%
                },
        columns/x/.style={
            column name={},
            postproc cell content/.code={}
        }
    }
}
\pgfplotsset{compat=newest}

\newcommand{\seq}{\mathit{Seq}\xspace}
\newcommand{\activation}{\mathit{Activ}\xspace}

\newcommand{\order}{\mathit{order}\xspace}
\newcommand{\image}{\mathcal{I}\xspace}

\newcommand{\neuralnetwork}{\Xi\xspace}
\newcommand{\nbimages}{\mathit{NbIm}\xspace}
\newcommand{\dataset}{\mathcal{DS}\xspace}
\bibpunct[, ]{[}{]}{,}{n}{,}{,}%
\renewcommand\bibfont{\fontsize{10}{12}\selectfont}%
\makeatletter%
\def\NAT@def@citea{\def\@citea{\NAT@separator}}%
\makeatother%

\begin{document}

\begin{frontmatter}

\title{Found in Translation: semantic approaches for enhancing\\ AI interpretability in face verification}

\author[1,2,6]{Miriam Doh}
\ead{miriam.doh@umons.ac.be}
\author[4,5,3]{Caroline Mazini Rodrigues}
\ead{caroline.mazinirodrigues@esiee.fr}
\author[4]{Nicolas Boutry}
\author[5]{Laurent Najman}
\author[1]{Matei Mancas}
\author[1]{Bernard Gosselin}

\cortext[3]{Corresponding author.}
\cortext[6]{This work was supported by the ARIAC project (No. 2010235), funded by the Service Public de Wallonie (SPW Recherche).}
\affiliation[1]{organization={ISIA lab - Université de Mons (UMONS)},%
    city={Mons},
    country={Belgium}}

\affiliation[2]{organization={IRIDIA lab - Université Libre de Bruxelles (ULB)},%
    city={Brussels},
    country={Belgium}}

\affiliation[4]{organization={Laboratoire de Recherche de l'EPITA -- LRE},%
    addressline={14-16, Rue Voltaire}, 
    city={Le Kremlin-Bicêtre},
    postcode={94270}, 
    country={France}}
\affiliation[5]{organization={Univ Gustave Eiffel, CNRS, LIGM},%
    city={Marne-la-Vallée},
    postcode={77454}, 
    country={France}}

\begin{abstract}
The increasing complexity of machine learning models in computer vision, particularly in face verification, requires the development of explainable artificial intelligence (XAI) to enhance interpretability and transparency. This study extends previous work by integrating semantic concepts derived from human cognitive processes into XAI frameworks to bridge the comprehension gap between model outputs and human understanding. We propose a novel approach combining global and local explanations, using semantic features defined by user-selected facial landmarks to generate similarity maps and textual explanations via large language models (LLMs). The methodology was validated through quantitative experiments and user feedback, demonstrating improved interpretability. Results indicate that our semantic-based approach, particularly the most detailed set, offers a more nuanced understanding of model decisions than traditional methods. User studies highlight a preference for our semantic explanations over traditional pixel-based heatmaps, emphasizing the benefits of human-centric interpretability in AI. This work contributes to the ongoing efforts to create XAI frameworks that align AI models behaviour with human cognitive processes, fostering trust and acceptance in critical applications.
\end{abstract}


\end{frontmatter}

\section{Introduction}

The evolution and increasing complexity of machine learning models, particularly in computer vision, has highlighted the need for greater transparency and interpretability. This need has spurred the development of the field of Explainable AI (XAI) \cite{confalonieri2021historical,swartout1991explanations}, which aims to elucidate machine decision processes for human understanding. Accuracy, transparency, and interpretability are at the heart of XAI.

Although there is no officially agreed-upon terminology for ``interpretability,'' in this context we define it as the ability to present machine reasoning in terms that are understandable to humans, including a wide range of users, from beginners to experts. Such clarification is fundamental for building trust and promoting the acceptance of advanced technologies in critical sectors where automated decisions have significant implications \cite{cao2022reform,caruana2015intelligible}.

One important computer vision XAI technique that enhances interpretation is heatmap visualization. It provides an intuitive visual representation of the image areas that influence the most the model decisions. This approach has succeeded because it visually simplifies the machine's predictive logic interpretation. However, the effectiveness of such tools and other XAI techniques depend on their ability to meet users' specific needs and competencies. Previous research has indicated that heatmaps may only be intuitive for less experienced users \cite{kim2023help, jordan2015machine, machlev2022explainable}. The main issue with this approach is that the explanation is provided at the pixel level (low level) and for local explanations (individual images), leaving it to the user to align this response with their cognitive process -- \textit{i.e.}, map the individual/pixel-level explanation to a semantic/conceptual level that globally makes sense to the user.

In previous work \cite{Doh2024Beware}, inspired by human perceptual processes, an initial framework was proposed for facial verification. The framework introduces XAI solutions based on semantic concepts definable by a user to provide explanations of the decision-making process, understandable in human terms. This approach focused on identifying user-defined concepts --- such as eyes and nose---, to which the models are sensitive for most examples (concept extraction phase), allowing a global understanding of the model's behavior (global explanation). Then, knowing the models' preferences for specific concepts, a model-agnostic local method is used to provide explanations based on the obtained most relevant human concepts by analyzing their contribution in terms of similarities and dissimilarities (local explanation).

This work proposes an extension of the previous framework by \textbf{including a different strategy} for combining global concepts' importance intro local explanations for face verification task; \textbf{validating the initial approach} through further experiments to confirm its effectiveness; and \textbf{evaluating users' perception} of the proposed explanations.

Specifically, we quantitatively evaluate the methods used in the concept extraction process by comparing context-aware (LIME~\cite{25ribeiro2016should} and MAGE~\cite{mazini:2024:infosciences}) and local XAI techniques (KernelSHAP~\cite{lundberg2017nips}). In these experiments, we evaluate the model explanations' fidelity. For the qualitative evaluation, we analyze the users' preferences according to different face segmentations, and text explanations. 

The main contributions of this work focus on:
\begin{itemize}
    \item Proposing a novel strategy to combine global and local explanations for face verification task.
    \item Adapting context-aware algorithms for extracting models' global concepts (LIME~\cite{25ribeiro2016should} and MAGE~\cite{mazini:2024:infosciences}).
    \item Quantitatively evaluating concept extraction from models through concepts' occlusion experiments.
    \item Integrating textual explanations through Large Language Models (LLMs).
    \item Conducting a comprehensive evaluation of the impact of these techniques on human perception of AI explanations.
\end{itemize}

Through these efforts, we aim to bridge the interpretive gap between humans and machines, advancing the creation of XAI frameworks that align the model's artificial cognition with the human cognition. Section~\ref{ch:sota} presents the state-of-the-art methods and our motivation; Section~\ref{class} deals with our methodology; Section~\ref{ch:metrics} provides the evaluation metrics; Section~\ref{ch:exp} shows the experiments, results and discussion, and Section~\ref{ch:conclusion} concludes the work and provides future work tracks.

\section{Related works and motivation}
\label{ch:sota}

In recent years, significant efforts have been made in computer vision to develop methods that explain model predictions using attribution techniques. Specifically, attribution techniques aim to identify which parts of the input data (such as image regions, pixels, or features) most influenced the model's decision, thus providing insights into how models make predictions. Those efforts include Class images Activation Mapping (CAM) \cite{38zhou2016learning,28selvaraju2017grad} and Layer-wise Relevance Propagation (LRP) \cite{2binder2016layer}, which are model-specific, as well as model-agnostic methods such as LIME \cite{25ribeiro2016should} and RISE \cite{23petsiuk2018rise}.

As defined by \cite{renftle2024algorithms}, \textit{``Given the core question Q* addressed by XAI algorithms, [...] one of the challenges is `Translation'. The [...] challenge in answering Q* is to translate the technical attributes that ML models use to discriminate between data items into interpreted attributes."} This translation, which we associate with the concept of ``interpretability,'' is important in explainability.

Despite their popularity in the research community, attribution methods do not always align with this essential principle of explainability. One of the main challenges in this field is translating XAI solutions for a less experienced audience (layperson), as these methods provide information at the pixel level, leaving the cognitive burden of interpretation to the users \cite{apicella2020middle}.

Several works have attempted to bridge this gap by translating granular information into ``concepts" that are semantically meaningful to humans. One trend in this direction is to use global explanations, \textit{i.e.}, explanations related to an entire class of the trained neural classifier, involving human-understandable concepts to explain attributes or abstractions. Examples include Prototype \cite{kim2018interpretability} and Concept Activation Vector (CAV) \cite{ghorbani2019towards}, which aim to visually represent the internal states of the neural network associated with a specific class. CAVs should represent concepts understandable to humans.

However, the use cases proposed to demonstrate how these approaches explain their decisions are often limited and assume prior knowledge of the discriminative concepts for the task. Moreover, most datasets of human concepts may not be available for a specific domain and must be collected at high costs. Even if the dataset is available, there is a considerable risk that user-defined concepts may be incomplete or inaccurate, leading to poor or biased explanations \cite{LONGO2024102301,10136140}.

One promising approach is described in \cite{rodrigues2024unsupervised} and proposes a method to discover interpretable visual concepts without supervision. Using clustering techniques, this method identifies features that users can easily understand, generating interpretable visual concepts directly from the data without requiring a predefined dataset of human concepts.

Another work seeking to use ``concepts" larger than pixels and meaningful to humans from a local explanation perspective is that of Apicella et al. \cite{apicella2022exploiting}. They recently proposed a method based on hierarchical image segmentation using autoencoders, providing local explanations in various forms for the same ML model. This method offers greater granularity in explanations, better adaptation to the different competence levels of users. The LIME method, based on superpixels, can also be considered among these approaches.

Building on these premises, the core idea of our previous paper \cite{Doh2024Beware} was to combine middle-level spatial explanations with concepts, merging the two foundations that have driven the works mentioned above.

Specifically, we aimed to use ``superpixels" explicitly linked to semantically meaningful, user-definable concepts. This ``personalized" approach integrates spatial context with human-understandable semantics, enabling users to define the concepts and thus providing spatially coherent and semantically relevant explanations. 

Moreover, by leveraging theories from cognitive psychology, we believe our approach gains a significant advantage, aligning with what has been coined as XAI 2.0 \cite{LONGO2024102301}. This new paradigm in XAI promotes a broader, multidisciplinary approach that integrates insights from various fields to develop more effective and comprehensive XAI solutions.
Following this approach, our work focuses on face verification. In this paper, we aim to quantitatively evaluate this framework and test it experimentally, including user feedback, addressing the need for more empirical studies on the effectiveness of explanations.

\section{Methodology}
\label{class}

\begin{figure*}[ht]
  \centering
  \includegraphics[width=0.95\textwidth]{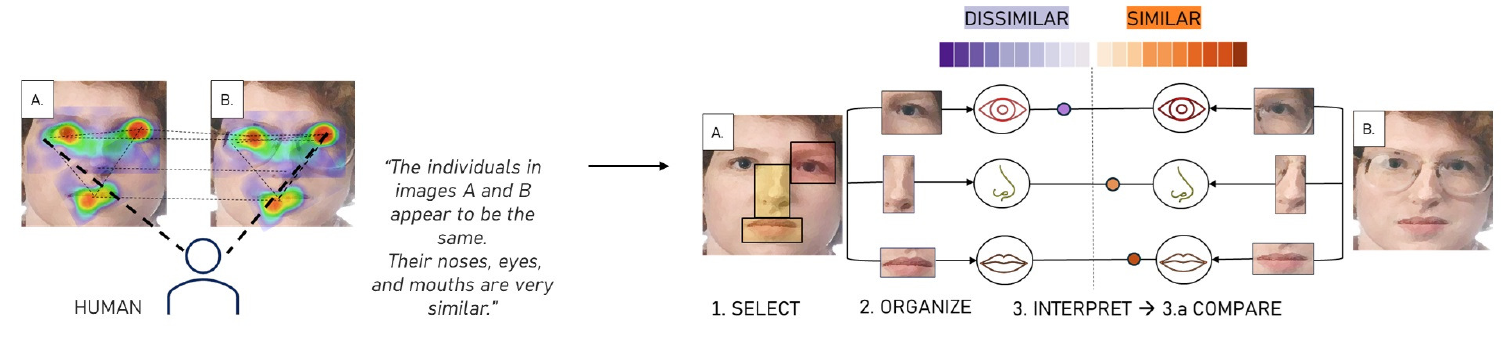}
  \caption{On the left, an illustration shows how humans perform face recognition by focusing on specific facial areas. On the right, we present an adaptation of the XAI Perceptual Processing Framework, originally proposed by Zhang et al. \cite{Zhang2022}, specifically tailored for face verification, drawing inspiration from how humans process visual stimuli.}
  \label{fig:first_paper_idea}
\end{figure*}

In our previous work, we explored how inspiration from human cognitive processes can enhance the understanding of decisions made by AI systems, specifically focusing on facial recognition. Through cognitive psychology, we highlighted how specific facial areas, such as the eyes and nose, play a crucial role in perceiving and recognizing faces. According to human perceptual processes, these visual stimuli are organized into meaningful concepts through stages of selection, organization, and interpretation, facilitating the categorization of faces. An emblematic example of this process is the recognition of specific resemblances, as expressed in statements like: \textit{You look like your mother. You have the same eyes''}, where the comparison of eyes'' underscores a similarity through a well-defined semantic area (Figure \ref{fig:first_paper_idea}).

\begin{figure*}[!ht]
    \centering
    \includegraphics[width=\linewidth]{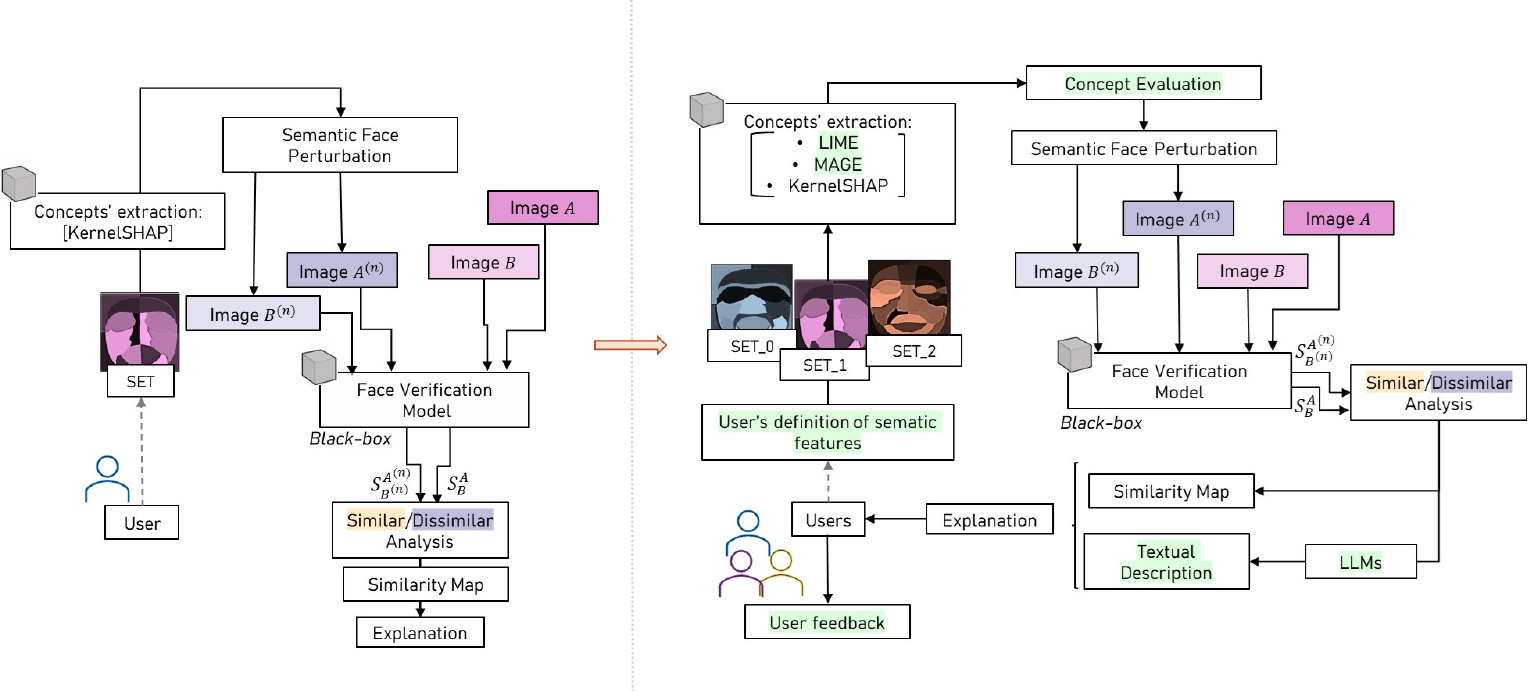}
    \caption{This figure contrasts the previously established framework (left) with the newly proposed framework presented in this paper (right). The new framework includes several additional components, highlighted in green. Specifically, it introduces three hypothetical semantic sets to evaluate the variability of the proposed method. Moreover, the new framework incorporates the evaluation of three concept extraction methods (KernelSHAP, MAGE, LIME), whereas the previous work utilized only KernelSHAP without evaluation. The explanation visualization has been expanded to include textual descriptions via LLM models. Additionally, user feedback has been incorporated, which was not collected in the prior work.}
    \label{fig:frame}
\end{figure*}

Building on this idea, we raised concerns about the effectiveness of traditional saliency maps in computer vision, questioning their ability to emulate human reasoning processes. In particular, human saliency maps will focus on specific face parts (based on top-down knowledge about faces and bottom-up rare features). At the same time, AI model saliency maps may also focus on textures or other features humans do not consider. In response to this critique, we developed a flowchart incorporating principles derived from cognitive psychology to refine the interpretation of AI decisions through an input perturbation approach, as shown in the initial schema in Figure~\ref{fig:frame}.

The core of our approach is to explain a face verification system in an agnostic way (black box) based on a cosine similarity score, $S^{A}_B$, specifically defined as $S^{A}_B = \frac{\mathbf{f_A} \cdot \mathbf{f_B}}{||\mathbf{f_A}||\,||\mathbf{f_B}||}$, where f\textsubscript{A} and f\textsubscript{B} are the feature vectors extracted by the model. This semantic perturbation approach aims to identify which facial areas are perceived as similar or dissimilar by modifying images $A$ and $B$ to produce new versions, $A_{(n)}$ and $B_{(n)}$, from which specific semantic areas identified as $n$ have been removed. By comparing the new similarity score, $S^{A_(n)}_{B_(n)}$, with the original score, we determine the importance of the removed areas in the perception of similarity.

We emphasize that this masking process is applied only to globally relevant semantic areas identified as the model's key concepts rather than to all hypothetically user-defined concepts. By evaluating the change in score, $\Delta s$, we can infer the contribution of the removed parts to the perceived similarity. A decrease in $S^{A_(n)}_{B_(n)}$ compared to $S^{A}_B$ implies a positive contribution of the excluded areas to similarity ($\Delta s\geq 0$), while an increase suggests a negative contribution ($\Delta s<0$).
Expanding upon this framework, the current study proposes an evolution and improvement of this flowchart. It intends to validate the various stages of the process through user perception and quantitative experiments to thoroughly investigate the initially outlined process.
In the subsequent section, we will explore each phase of the new flowchart (second flowchart in Figure~\ref{fig:frame}) in detail and its modifications. We explain the process in four phases: \textit{1) semantic features} definition, \textit{2) concepts extraction}, \textit{3) similarity map}, and \textit{ 4) transcription of semantic contributions using LLMs}.

\subsection{Definition of Semantic Features}

\begin{figure}[H]
\centering
\includegraphics[width=0.7\linewidth]{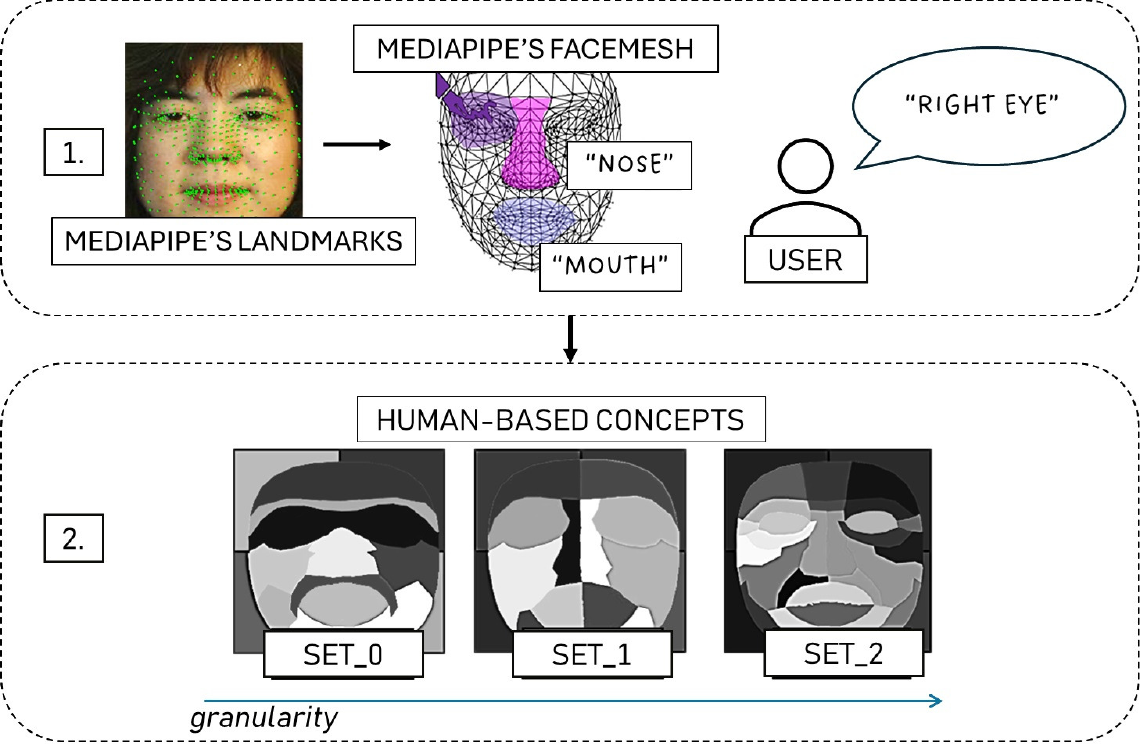}
\caption{The process of creating human-based semantic features. (1) Mediapipe's landmarks projected onto an example face. Using Mediapipe's facemesh, users can define semantic areas by selecting specific landmarks, as shown in the example. (2) Three sets of human-based concepts with varying granularity (SET\_0, SET\_1, SET\_2) created from these user-defined areas. SET\_0 and SET\_1 have 13 features each, while SET\_2 has 30 features.}
\label{fig:human_concepts}
\end{figure}

Expanding on the suggestions from previous work to bridge the comprehension gap between users and the explanations provided by models, we have utilized the landmarks provided by Mediapipe \cite{lugaresi2019mediapipe}, an open-source framework by Google, as a foundation for creating customizable semantic maps. In our earlier study, we identified 13 hypothetical semantic areas of the face, including background zones, where "background" refers to any area outside the facial landmarks. This study explores various semantic configurations by varying the number of concepts for two primary reasons, as shown in Figure~\ref{fig:human_concepts}
.

First, employing different semantic masks allows us to assess the method's sensitivity to variations in these masks. Second, we aim to avoid imposing a single semantic definition of facial areas while recognizing that collecting multiple definitions would be prohibitively time-consuming for calculating the global importance of concepts. Therefore, we defined three sets representing different levels of detail, ranging from SET\_0 (the least detailed with 13 broad features, where facial regions like the eyes, nose, and mouth are treated as single unified areas without separation between left and right sides) and SET\_1  (also with 13 features, but with more granular division, including a distinction between the left and right sides of symmetrical facial features such as eyes and lips) to SET\_2  (the most refined with 30 features, providing even finer segmentation across various facial areas).

Figure~\ref{fig:human_concepts} illustrates the process of creating these semantic features. In step 1, Mediapipe's landmarks are projected onto an example face. Then, users define semantic areas by selecting specific landmarks on Mediapipe's facemesh. Step 2 shows the three sets of human-based concepts with varying levels of granularity that we hypothesized for this study.

The ability to define human concepts upon which to base the algorithm can bridge the comprehension gap by integrating user-defined semantics into the explanation. This facilitates a transition from explanations based on low-level features (individual pixels) to more comprehensible explanations based on middle-level features  (aggregates of pixels).

\subsection{Concepts Extraction}

Human cognition intuitively segments images into distinct semantic entities, such as eyes, nose, and mouth. Machines do not have this innate ability; the identification and differentiation of such concepts must be acquired through the training process. A significant challenge arises when the concepts learned by machines do not match those understood by humans, complicating the interpretation of machine decision processes.

Traditionally, XAI methods have focused on providing case-by-case explanations that elucidate the rationale behind the model's decisions for individual images. Previous work~\cite{Doh2024Beware} has introduced a methodology that aggregates these individual explanations into a comprehensive ranking of human-understandable concepts, prioritized by their importance to the model's inference process.

In this work, we compare three explanation methods used to find globally important semantic concepts. We use the framework proposed by~\cite{Doh2024Beware} but instead of only using KernelSHAP~\cite{lundberg2017nips}, which is a local explanation technique, we also compare it to the results of using two global-aware techniques: LIME~\cite{25ribeiro2016should} and MAGE~\cite{mazini:2024:infosciences}. With this evaluation, we want to verify if global-aware methods outperform local ones in the face verification tasks. In the sequel, we describe the three compared XAI techniques and the adaptations to the face verification problem.

\textbf{MAGE and EaOC:} For the first one, we adapt the MAGE technique proposed by Rodrigues~\textit{et.~al.}~\cite{mazini:2024:infosciences}. This method uses the last convolutional layer of the network to find groups of similar responses to input patterns. This is made by a decomposition of this layer response according to the behavior of each feature map dimension.

\begin{figure}[!ht]
    \centering
    \includegraphics[width=\linewidth]{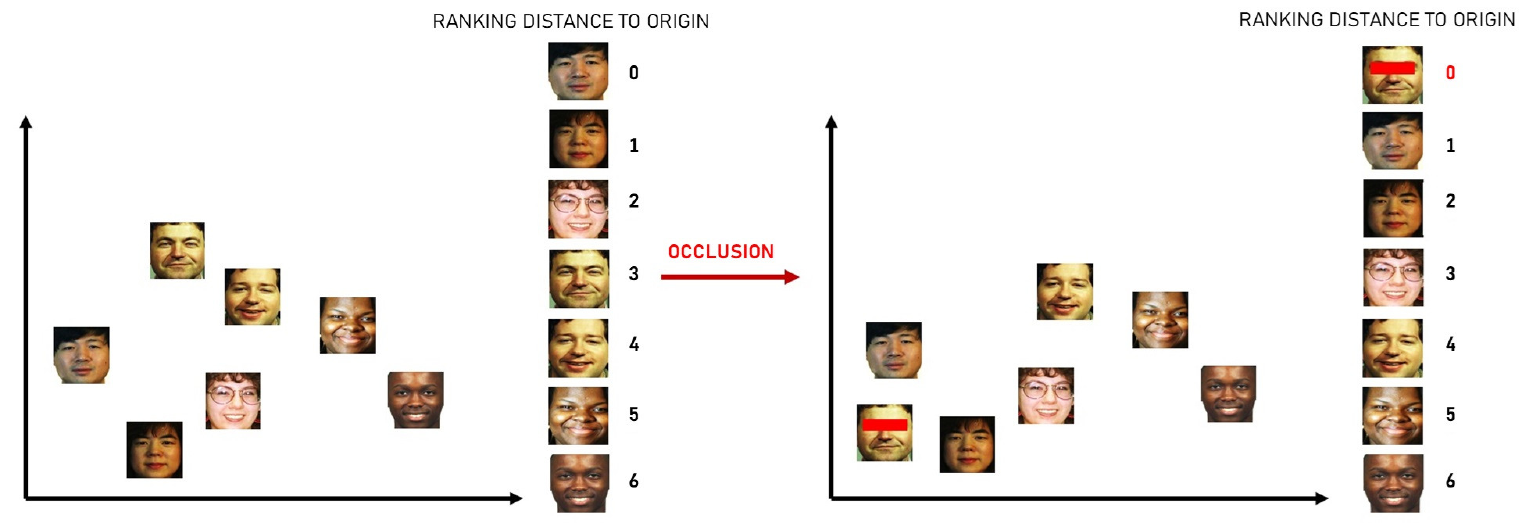}
    \caption{Illustrative example of EaOC behavior under an occlusion. Consider each represented image as the corresponding embedding for this image obtained by a trained model. Initially, given a set of images, we order the images according to their distance to the origin. After each occlusion, we calculate the orders again. The occluded image may change its order if the occlusion is impactful.}
    \label{fig:mage}
\end{figure}

Originally, after finding the Maximum Activation Groups (MAGs), clusters of different behaviors representing each one a concept, the authors proposed a visualization technique called Ms-IV~\cite{mazini:2024:infosciences} to show, in multiscale, the most important image regions in some image examples.

For the visualization Ms-IV, the evaluation of each image patch is made by occlusions. However, instead of using a distance metric to evaluate the difference before and after a patch occlusion, the authors propose the metric Class-aware Order Correlation (CaOC). This metric considers how the output space of a specific class changes after an occlusion, using rankings. 

The use of this methodology, MAGE, provides global-aware explanations for two reasons: firstly, it considers the activation patterns to obtain global groups of concepts learned by the model; secondly, it uses these \textit{global concepts} combined to an \textit{output space analysis} subjected to performed occlusions in the input.

Here, we propose to adapt Ms-IV to a human-driven segmentation visualization. We use the face segments (three different human concept segmentations presented in Figure~\ref{fig:human_concepts}) to replace the patches of the original method. Moreover, we adapt CaOC to, instead of analyzing the class space changes, we account for the complete embedding space. We will call the adapted CaOC from now on, \textit{Embedding-aware Order Correlation} (EaOC).

Let us consider a trained model $\neuralnetwork$ and a dataset $\dataset$ of  $\nbimages(\dataset)$ face images $(\image_l)_{l \in [1, \nbimages(\dataset)]}$. We denote as $\activation(\image_{l};\neuralnetwork)$ the embedding of image $\image_l$ obtained by the model $\neuralnetwork$. To obtain information on the model's output space, we calculate the Euclidean distance of all dataset images to the origin (norm-2) according to Equation~\ref{eq:output_space}:

\begin{equation}
    \mathit{OS}_{\activation}(\dataset) =  (||\activation\left(\image_{l};\neuralnetwork)\right||_2)_{l \in [1,\nbimages(\dataset)]}.
    \label{eq:output_space}
\end{equation}

As we have numeric values representing each image as distances to the space origin, we can also impose an order of images:

\begin{equation}
    \seq_{orig} = \order\left( \mathit{OS}_{\activation}(\dataset),\mathit{decreasing}\right)
    \label{eq:original_order}
\end{equation} 

which shows, from the most distant to the least distant, the positions of each image in a ranking of distances.

The idea here is to evaluate how occlusions of human-based concepts impact this spatial organization (calculated in Equation~\ref{eq:original_order}). Therefore, after an occlusion of the human-based concept $i$ with $i \in [1,\mathit{num}_\mathit{concepts}]$ on image $\image_j$ that is the $j^{th}$ image on $\seq_{orig}$, we can recalculate the order as follows:

\begin{equation}
    \seq_{occ}^{\dataset_j^{(i)}} = \order\left( \mathit{OS}_{\activation}(\dataset^{\dataset_j^{(i)}}),\mathit{decreasing}\right).
     \label{eq:occlusion_order}
\end{equation} 

On $\seq_{occ}^{\dataset_j^{(i)}}$ (Equation~\ref{eq:occlusion_order}) the image $\image_j$ might have a new order according to the impact of occlusion on its embedding. EaOC evaluates the difference between the original $\image_j$ order and the new one, conditioned to the occlusion of concept $i$: 

\begin{equation}
   \mathit{EaOC}(\image_j,i) = |\seq_{orig,j} - \seq_{occ,j}^{\dataset_j^{(i)}}|
    \label{eq:eaoc}
\end{equation} 

With Equation~\ref{eq:eaoc} we evaluate the concept's impact for this specific image. Figure~\ref{fig:mage} presents an illustrative example of the embedding space and the relation of images to obtain $\mathit{EaOC}(\image_j,i)$.

\textbf{LIME:} In addition to this adapted method, we also consider using Local Interpretable model-agnostic explanations (LIME) as a context-aware method. Even being local, the idea of approximating a local partition of the embedding space with a surrogate model induces spatial awareness. 

As the original method presents approximations to one specific class, we adapted the surrogate model to approximate the norm-1 of the embedding of each image.

\textbf{KernelSHAP:} As used in Doh~\textit{et.~al}~\cite{Doh2024Beware}, we also test KernelSHAP for concepts extraction. However, as a local explanation method, it represents the most important concepts for each image, that can be posteriorly combined. This method uses LIME~\cite{25ribeiro2016should}'s interpretable components with Shapley values~\cite{castro2009cor} to find the feature contribution to the model's output. 

\subsubsection{Concepts aggregation}
\label{sec:concepts_agg}

\begin{figure}[!ht]
    \centering
    \includegraphics[width=0.6\linewidth]{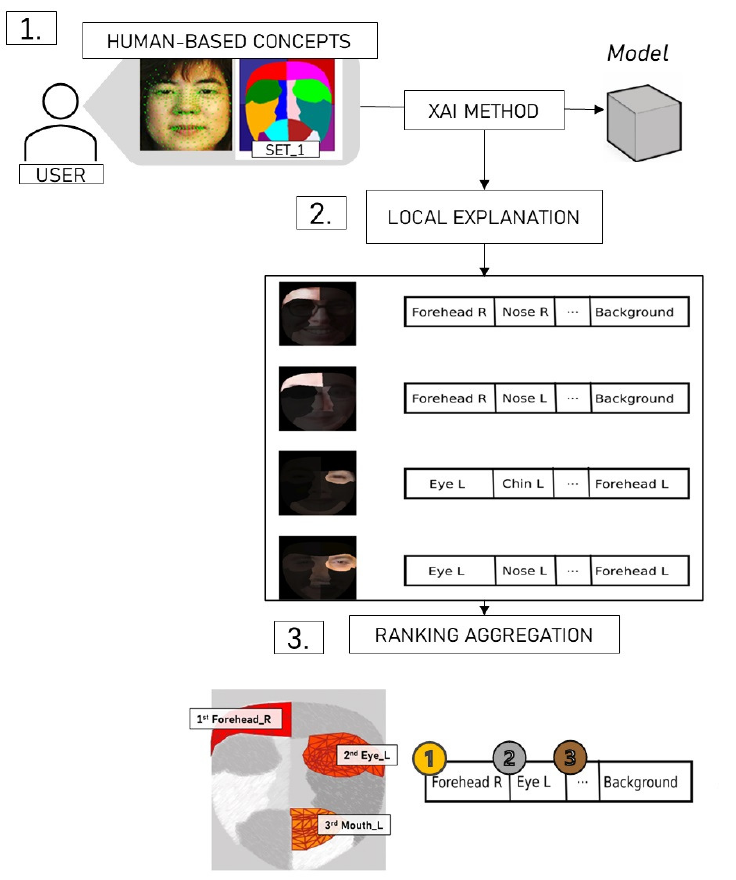}
    \caption{\textbf{Methodology to extract globally important concepts.} We use the human-segmented regions to obtain the explanations (using xAI methods such as LIME, KernelSHAP and MAGE) for all images (\textbf{1}). We order the segmented regions' importance, for each image, according to the xAI method \textbf{(2)}. Finally, we combine the orders into a final ranking that shows the most important face segments globally (\textbf{3}).}
    \label{fig:borda}
\end{figure}

At the end of our framework, each image explanation, using the segments from the human-based concepts' segmentation (as shown in Figure~\ref{fig:borda}~\textbf{(1)}), will have scores representing the occlusion impact of each segment for the face verification task. They can be used to visually find the most impactful facial segments. However, this will be local to this image explanation.

The idea behind \textit{concepts aggregation} is to find globally important segments, \textit{i.e.}, segments that impact most of the images, before proposing the visualization. We maintain this step as it is in the previous work, which consists of combining a set of image explanations before obtaining the mentioned occlusion scores. 
The difference is that we propose a comparison between the local XAI method KernelSHAP used previously, with global-aware XAI explanation methods (such as LIME and MAGE). We apply the XAI techniques in a set of images, with the human-based concepts' segmentation (Figures~\ref{fig:borda}~\textbf{(1)} and \textbf{(2)}), to order the most important regions for each image.

This order is a ranking of the most important parts, which we can aggregate to obtain the final most important facial segments (Figure~\ref{fig:borda}~\textbf{(3)}). A diversity of ranking aggregation method can be used, we opted to use BORDA count~\cite{borda1781}.

We proceed differently for MAGE, which already decomposes the network into different concepts. We first use the BORDA count to aggregate the concepts' ranking for each part of the network for the same image. Then, we do the same as previously described, we aggregate the ranking for different images. 

Previously, after we had these most important segments, we highlighted only the top most important segments for the images' visualization. However, in this work, instead of filtering the top concepts, we include the position of each segment in this global ranking as weights to influence the local explanations when visualizing single images. %

\subsection{Similarity map}
We adopted the approach described in~\cite{Doh2024Beware} to create similarity maps to provide local explanations, specifically implementing the single removal algorithm $S0$ initially introduced by \cite{mery2022black} with Gaussian masks and later adapted to semantic masks. 
This new approach differs in the fact that we incorporate the importance of each semantic face region, as determined in the \textit{Concepts Aggregation} (Section~\ref{sec:concepts_agg}), to weight the perturbations in Algorithm~\ref{alg:single_removal_s0}.
Assume we have $s$ face regions and a vector $\mathbf{O}$ of size $s$ where $\mathbf{O}_n$ represents the order of region $n$ (with 0 being the most important and $s-1$ the least important). The weight $g_n = s - \mathbf{O}_n$ is then used to multiply the perturbation impact of $\Delta S$ for face region $n$. The basic idea of this algorithm is to perturb facial images to understand the impact of specific facial regions on the similarity score between two images. The main steps are outlined in Algorithm ~\ref{alg:single_removal_s0}.

\begin{algorithm}[!ht]
\caption{Single Removal - S0}
\label{alg:single_removal_s0}
\begin{algorithmic}[1]
\State Initialize contribution maps $H0$\textsubscript{A} and $H0$\textsubscript{B} to zero for both images $A$ and $B$.
\For{each semantic part $n$}
    \State Calculate the similarity difference $\Delta S = g_n \cdot (S_{AB} - S_{A'(n)B'(n)})$.
    \State Calculate the contribution $C_n = \Delta S \cdot W_{(A,B)n}$.
    \State Update the contribution map $H0(A,n) = C_n \cdot M\textsuperscript{(A,n)}$ and $H0_{(B,n)} = C_n \cdot M\textsuperscript{(B,n)}$.
\EndFor

\State Normalize the contributions:
\For{each semantic part $n$}
    \If{$H0_{(A,n)} \geq 0$}
        \State $H0^+\textsubscript{(A,n)} = \frac{H0_{(A,n)}}{\sum_{H0_{(A,m)} \geq 0} |H0_{(A,m)}|}$
    \Else
        \State $H0^-\textsubscript{(A,n)} = \frac{H0_{(A,n)}}{\sum_{H0_{(A,m)} < 0} |H0_{(A,m)}|}$
    \EndIf
\EndFor

\State Calculate the final similarity map $S0$\textsubscript{A}:
\State $S0\textsubscript{A} = \sum_{n} (H0^+\textsubscript{(A,n)}  + H0^-\textsubscript{(A,n)} ) \cdot M\textsuperscript{(A,n)}$

\State Repeat steps 3 and 4 to obtain $S0$\textsubscript{B} for image $B$.

\end{algorithmic}
\end{algorithm}
The algorithm begins by initializing the contribution maps $H0$\textsubscript{A} and $H0$\textsubscript{B} to zero for both images $A$ and $B$. For each semantic part $n$, the similarity difference $\Delta S$ is calculated between the original similarity score $S_{AB}$ and the new similarity score $S_{A'(n)B'(n)}$ obtained after removing part $n$. We include a weighting term $g_n$ to account for the importance of the semantic part as a global concept. The contribution $C_n$ is determined by multiplying $\Delta S$ by the relative weight $W_{(A,B)n}$. The contribution maps $H0_{(A,n)}$ and $H0_{(B,n)}$ are then updated with the values of the contributions multiplied by the semantic mask of part $n$.

Once all contributions are calculated, they are normalized separately for positive and negative contributions. If $H0_{(A,n)}$ is positive, it is normalized by dividing by the sum of positive contributions; if it is negative, it is normalized by dividing by the sum of negative contributions.

Finally, the overall similarity map $S0$\textsubscript{A} is obtained by summing the normalized contributions $H0^+$\textsubscript{(A,n)} and $H0^-$\textsubscript{(A,n)} for all semantic parts $n$. This process is repeated to obtain the final similarity map $S0$\textsubscript{B} for image $B$.
\subsection{Generation of Textual Explanations Using Large Language Models}
\label{sec:textrans}
In previous work, a similarity map and a table reporting the contributions of each top semantic area were produced. To make this output more accessible to a general audience, we included a textual transcription of the table using large language models (LLMs). Specifically, we employed three small LLMs available on Hugging Face: CodeLlama-7B \cite{rozière2024codellamaopenfoundation}, Zephyr-7B \cite{huggingface2023zephyr}, and Beagle14-7B \cite{huggingface2023beagle}. These models were selected for their ability to provide quick, coherent textual explanations without requiring training. We tested their performance within the LMstudio environment \cite{lmstudio2023}.

CodeLlama-7B, developed by Meta, is optimized for code-related tasks but also generates coherent text, making it suitable for explaining technical content. Zephyr-7B is designed for natural language understanding and generation across various domains. Beagle14-7B is known for fine-tuning instruction following and reasoning tasks, offering precise and contextually relevant responses.

We provided these models with a specific prompt to generate explanations of the table values \texttt{[contributions\_table]}. The prompt contextualized the task, explaining that a face verification system assigns a cosine similarity score \texttt{[cosine\_similarity\_percentage]} between two images and outlined how positive or negative values indicate similarity or dissimilarity in semantic areas. In particular, the models were instructed to explain the cosine similarity score by discussing the impact of each semantic area on the final result, using colour maps to indicate levels of similarity.

The actual prompt used in this process is included below:
\begin{justify}
\textbf{Prompt:} 

\texttt{"Context: A face verification system assigns a cosine similarity score between two images. In this instance, the cosine similarity is \\ \textcolor{teal}{[cosine\_similarity\_percentage]} \textcolor{darkgray}{(a percentage from 0 to 100\%)}. From the \\model's knowledge, several main human-understandable concepts are extracted; these concepts are used to explain the model’s output (cosine similarity). These concepts are associated with a similar/dissimilar score. Specifically, when a value is positive or equal to zero \textcolor{darkgray}{( $\geq$0)}, the model perceives these areas in the two images as similar. Conversely, they are seen as dissimilar when the value is negative (example: $-$0.5 ):\textcolor{teal}{[contributions\_table]}. Given that a color map is displayed where shades of purple indicate dissimilarity and shades of orange indicate similarity, with color intensity proportional to the magnitude of the similarity or dissimilarity, provide a simple \\ explanation of why the cosine similarity between the two images is \\ \textcolor{teal}{[cosine\_similarity\_percentage]}. \textcolor{darkgray}{No long explanation}"} 
\end{justify} 
The inclusion of specific instructions in the prompt—such as \textit{``a percentage from 0 to 100\%"}, \textit{``example: $\geq$ 0/-0.5"}, and \textit{``No long explanation"}—was designed to ensure that all three models generated clear, consistent, and accessible responses. These instructions provided a clear structure for creating concise and uniform explanations.

In \ref{sec:appendix_A}, we report the complete outputs of the examples shown in the following user survey, along with a brief explanation of why we incorporated these specific instructions into the prompt and the limitations of using LLMs for textual transcription in our task.

The use of these models enabled the generation of an additional layer of explanation, automatically producing textual descriptions that adapt in style depending on the model employed. This approach significantly enhances the interpretability of the model’s output, making it more accessible to a broader audience. Moreover, it provides a useful framework for assessing different explanation styles, allowing for a better understanding of user preferences and improving overall comprehension.
\section{Metrics}
\label{ch:metrics}

\subsection{Concept Evaluation Experiments}
Inspired by the metrics of Bommer~\textit{et~al.}~\cite{bommer2023corr}, we will test three aspects of the obtained models' concept explanations: \textit{1) faithfulness, 2) sensitivity} and \textit{3) randomization}. Faithfulness refers to how close an explanation is to the real model's behavior. Sensitivity measures the impact of input changes on the model's output. Finally, randomization analyzes the impact of random changes on the model's output.

\textbf{Occlusion of top concepts:} These experiments test faithfulness and sensitivity. After finding the model's global concepts from the concepts' aggregation (Section~\ref{sec:concepts_agg}), we iteratively occlude from the top concept to the least important concept. We expect, if we correctly choose the most important concepts, to have a high increasing rate of the difference (to the original output) at the beginning (top concepts) and a reduction of this rate at the end.

\textbf{Randomized occlusion:} Along with the occlusion of concepts, we perform the same experiment with a randomly generated order of concepts (to be occluded). We expect these results to behave as a low boundary to the other XAI methods.

\subsection{User Evaluation and Feedback Methodology}
\label{sec:userevaluation}
An additional aspect we sought to investigate was the users' subjective perception of not only the system's explanations but also the overall framework. Specifically, it was necessary to obtain a comprehensive set of evaluations from users. Given that many design choices within the framework were predicated on the assumption that they would enhance interpretability, assessing these choices through feedback from actual users was imperative.

We solicited feedback on the proposed semantic sets, particularly focusing on the clarity and utility of supplementary information accompanying the similarity map, such as tables or descriptive text generated via LLMs. Furthermore, we requested users to express their preference between traditional methods (LIME with superpixels) and our proposed semantic approach, specifically evaluating the clarity and user-friendliness of each method.

Our objective was to engage a highly diverse audience in terms of background and to thoroughly explore the issue of interpretability of explanations. To this end, the survey was disseminated via two social media platforms: Instagram (using a public account of one of the authors, which had over 1,000 followers) and Reddit, specifically in the \texttt{r/SampleSize} group, which comprises 222,000 members who voluntarily respond to surveys. Additionally, we utilized mailing lists from various laboratories with expertise in AI, law, and social studies at the universities of the authors.

\section{Experiments and Results}
\label{ch:exp}

The experiments utilized the color FERET database \cite{phillips1998feret}, wherein the images underwent preprocessing using the Multi-task Cascaded Convolutional Networks (MTCNN) \cite{16li2020face} technique to crop the faces to a uniform dimension (NxN, where N equals 256). The methods were evaluated on FaceNet \cite{schroff2015facenet} models trained using the CasiaWebFace \cite{yi2014learning}  and VGGFace2 \cite{massoli2020ivc} datasets.

\subsection{Semantic Extraction}

\begin{figure}[!ht]
    \centering
    \subfloat[]{\includegraphics[width=0.45\linewidth]{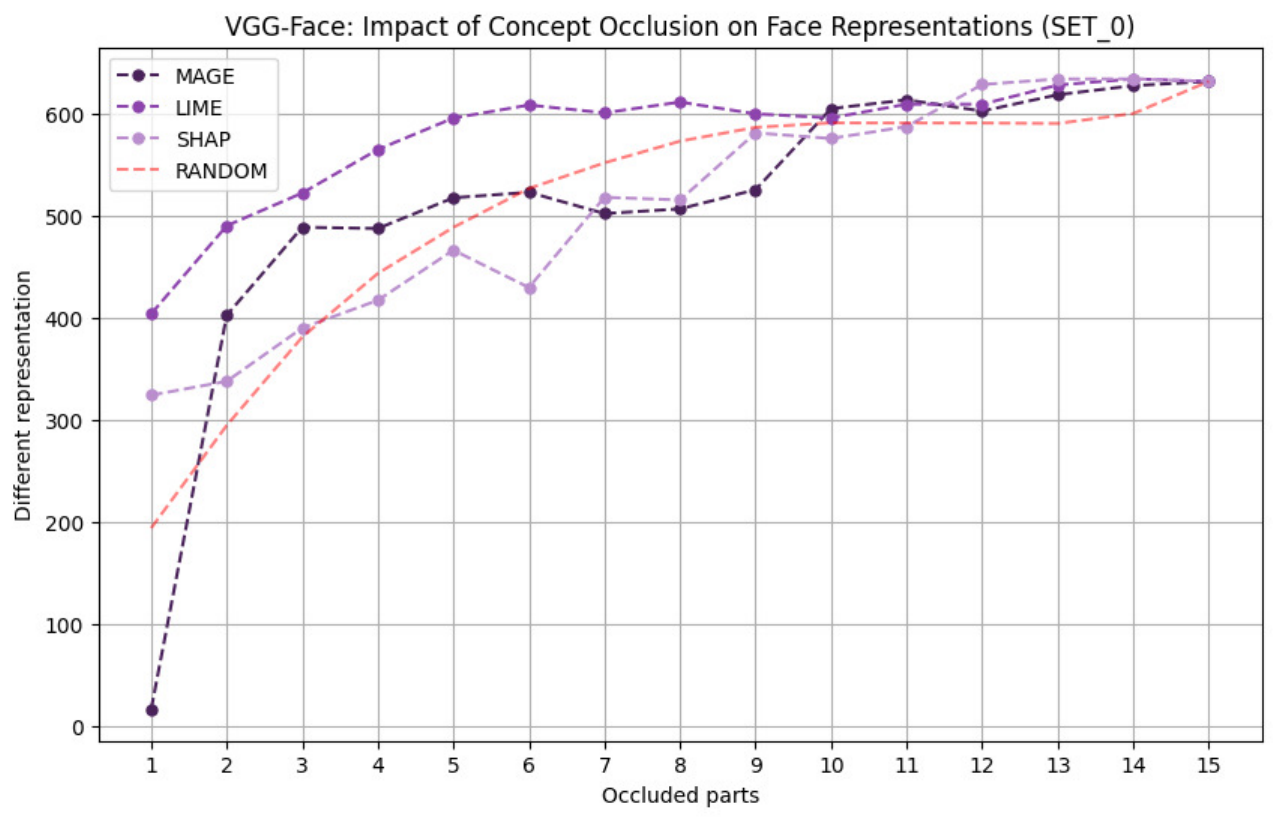} \label{fig:image1_rep}}
    ~
     \subfloat[]{\includegraphics[width=0.45\linewidth]{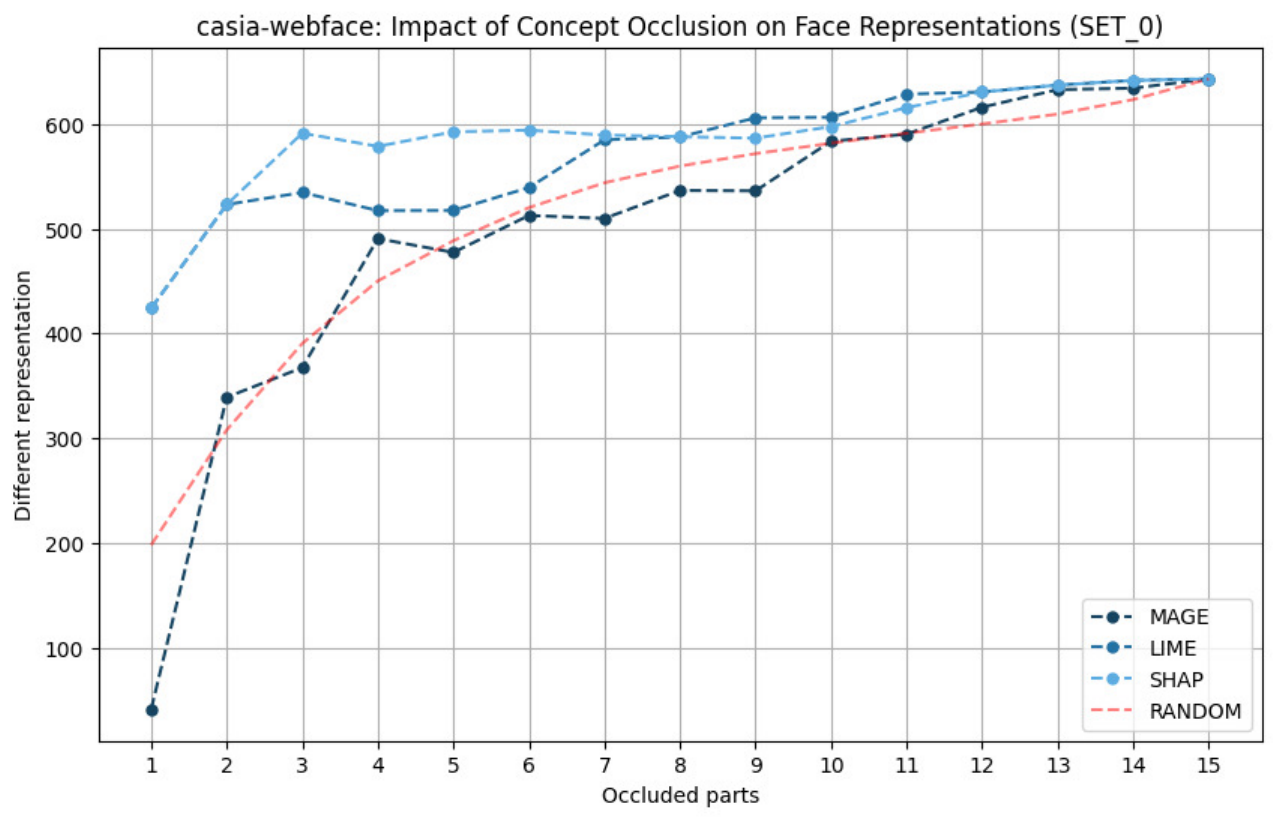} \label{fig:image4_rep}}
     \\
    \subfloat[]{\includegraphics[width=0.45\linewidth]{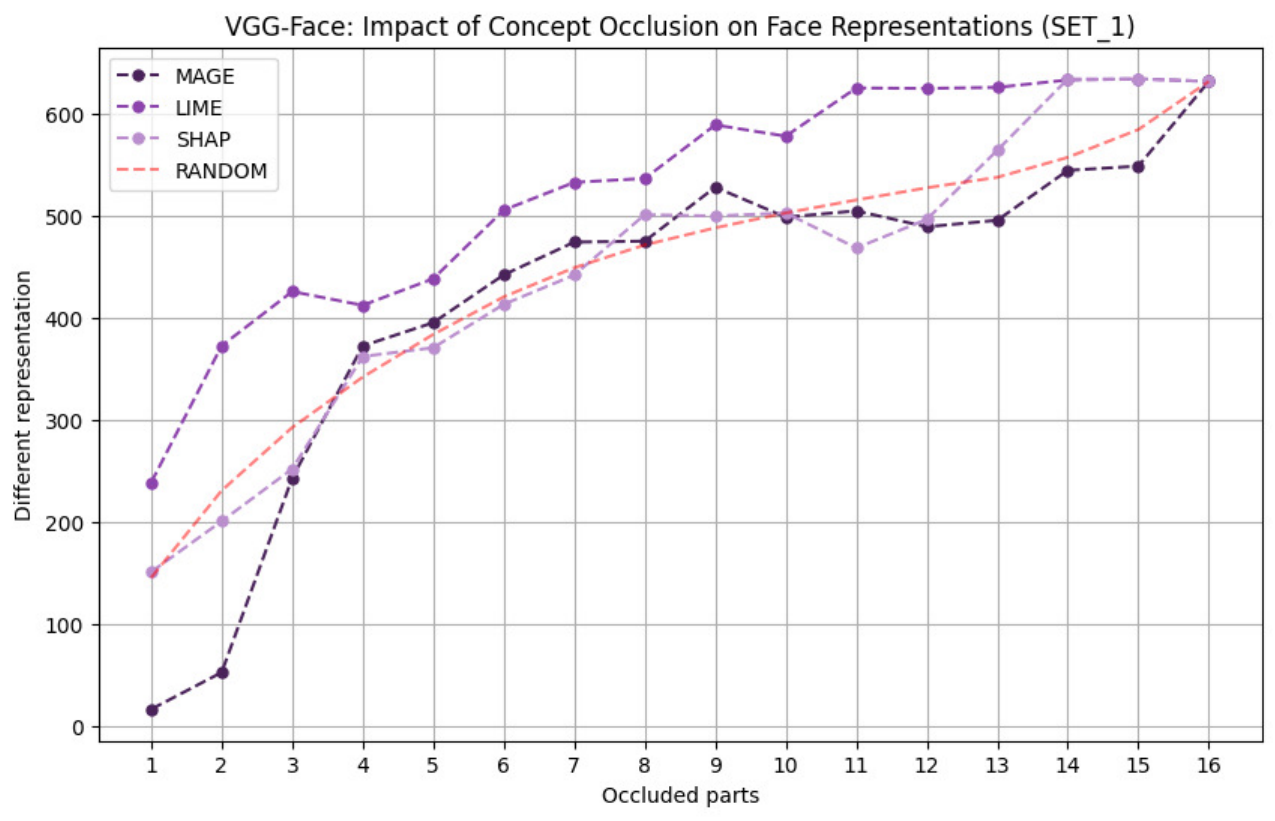} \label{fig:image2_rep}}
    ~
     \subfloat[]{\includegraphics[width=0.45\linewidth]{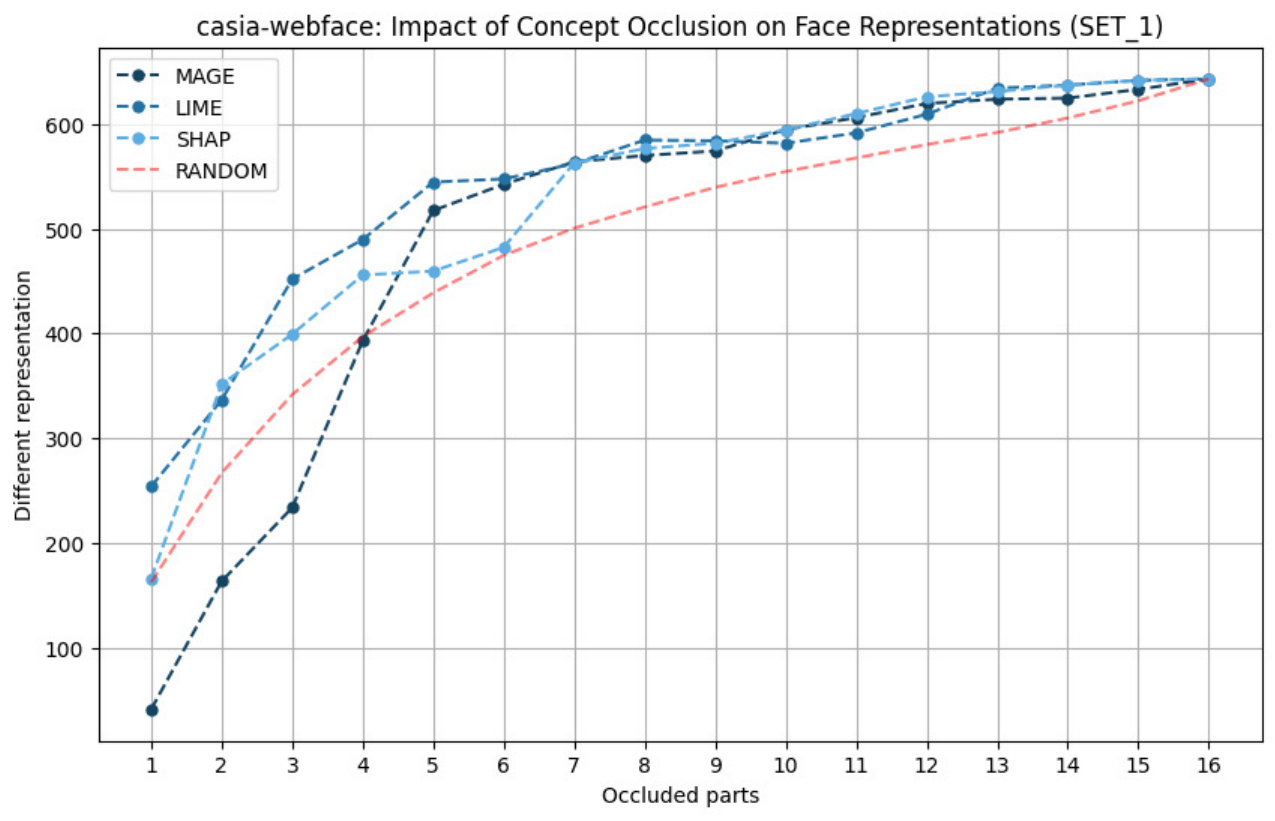} \label{fig:image5_rep}}
    \\
    \subfloat[]{\includegraphics[width=0.45\linewidth]{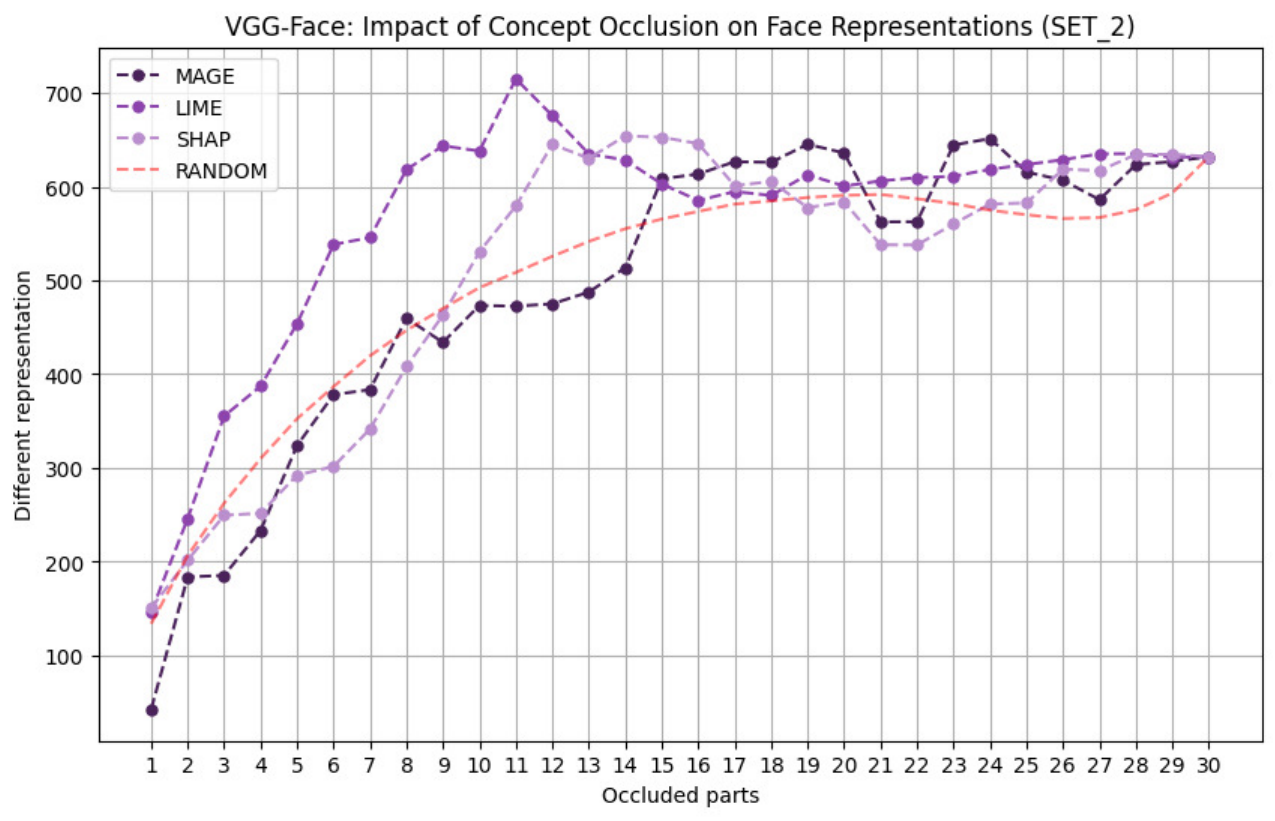} \label{fig:image3_rep}}
    ~
    \subfloat[]{\includegraphics[width=0.45\linewidth]{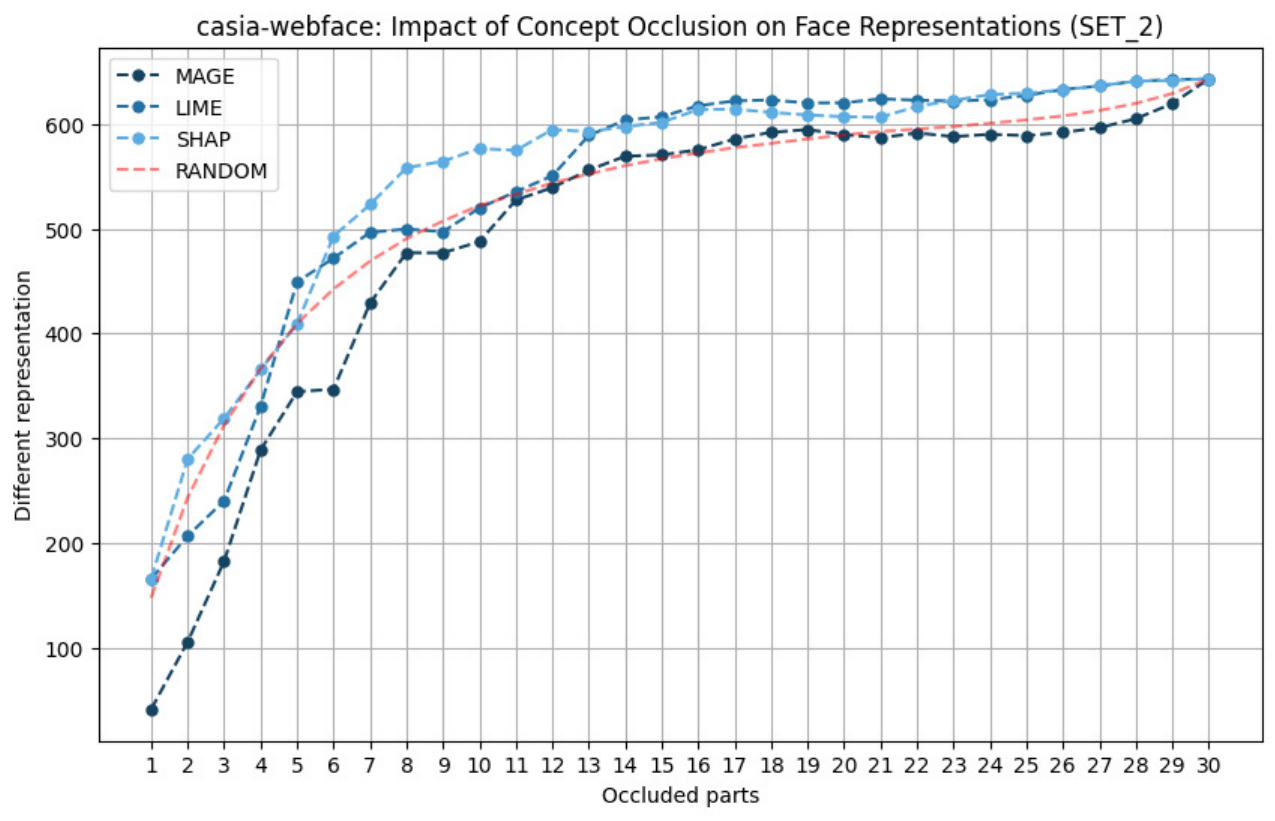} \label{fig:image6_rep}}
    \caption{Difference values of face representations by successive occluding from the most to the least important concepts found for three different semantic sets of face regions, for the VGGFace (a, c, and e) and Casia (b, d, and f) models. The x-axis presents the number of occluded parts, and the y-axis presents the Euclidean distance between the original and occluded face representation. We show the comparison of three XAI techniques to find concepts (MAGE, LIME, and SHAP) based on three semantic sets. We compare the results with a randomly based occlusion of the concepts for the three semantic sets. The random approach represents an average behavior for choosing concepts, and superior values represent good performances.}
    \label{fig:occ_representation_vgg}
\end{figure}

In these experiments, we analyze how much the top obtained concepts impact the final face representation and face verification task. We successively occlude image semantic regions representing the concepts, from most important to least important, and calculate the difference from the original output. We test three different semantic sets of regions. For the face representation comparison, we compare the difference between the 512-dimensional vectors of 750 images before and after occlusion using the Euclidean distance. We present the results for Casia-WebFace and VGGFace2 trained models in Figure~\ref{fig:occ_representation_vgg}.

\begin{figure}[!ht]
     \centering
     \subfloat[]{\includegraphics[width=0.45\linewidth]{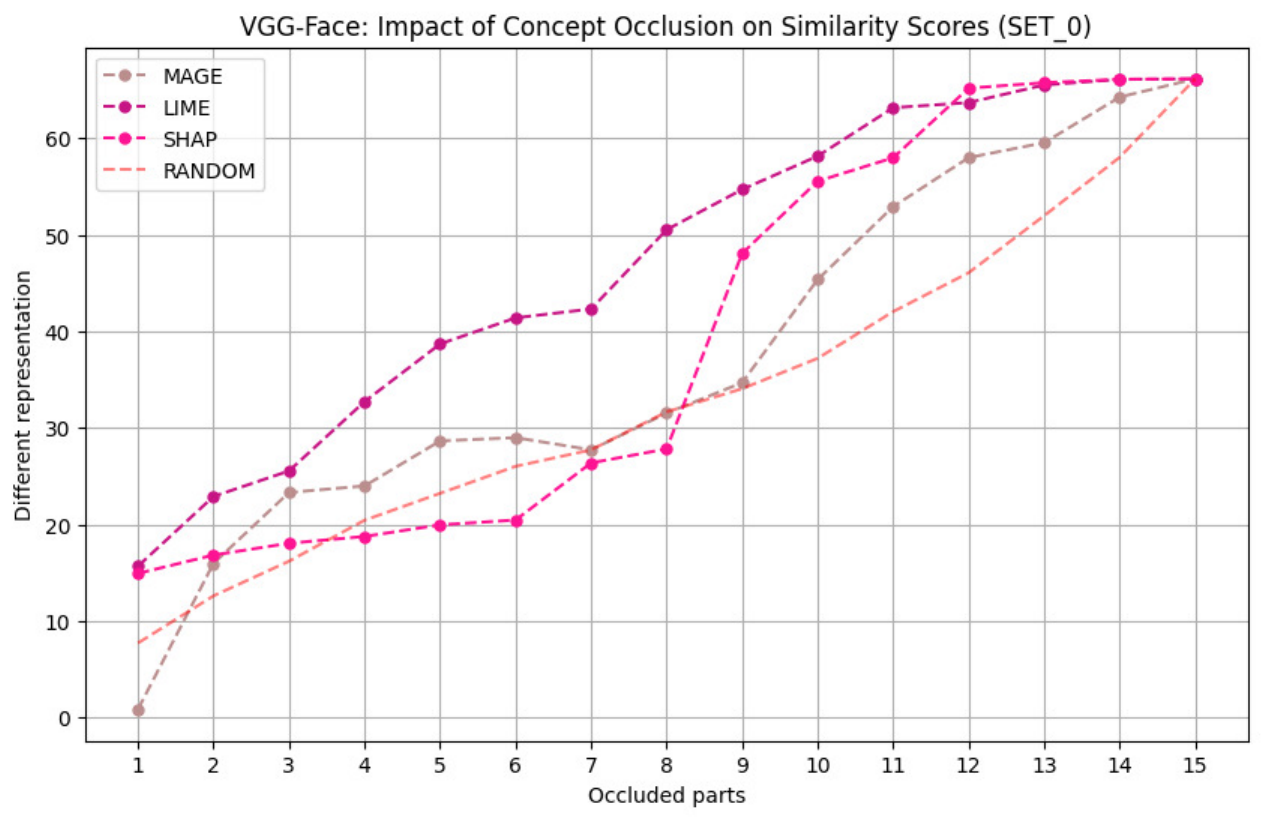} \label{fig:image1_sim}}
     ~ \subfloat[]{\includegraphics[width=0.45\linewidth]{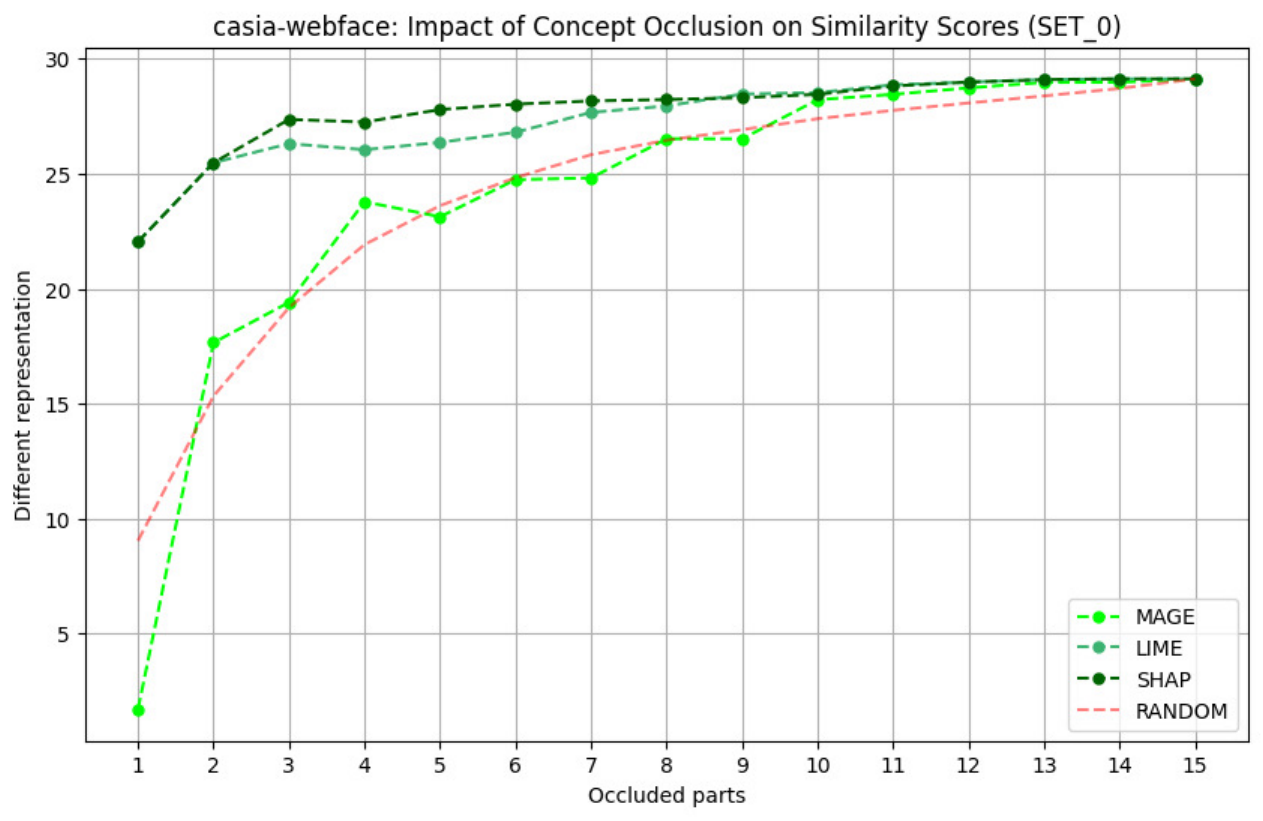} \label{fig:image4_sim}}
     \\
     \subfloat[]{\includegraphics[width=0.45\linewidth]{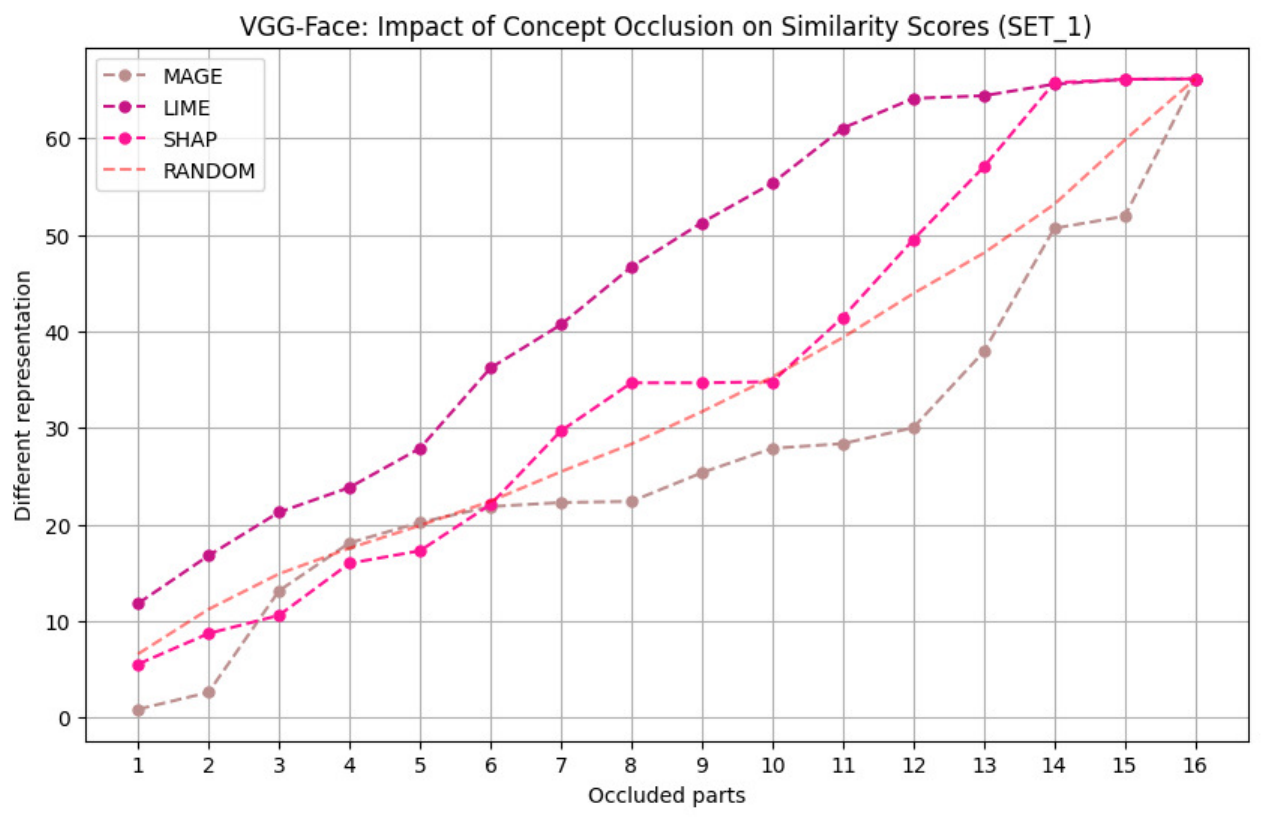} \label{fig:image2_sim}}
     ~ \subfloat[]{\includegraphics[width=0.45\linewidth]{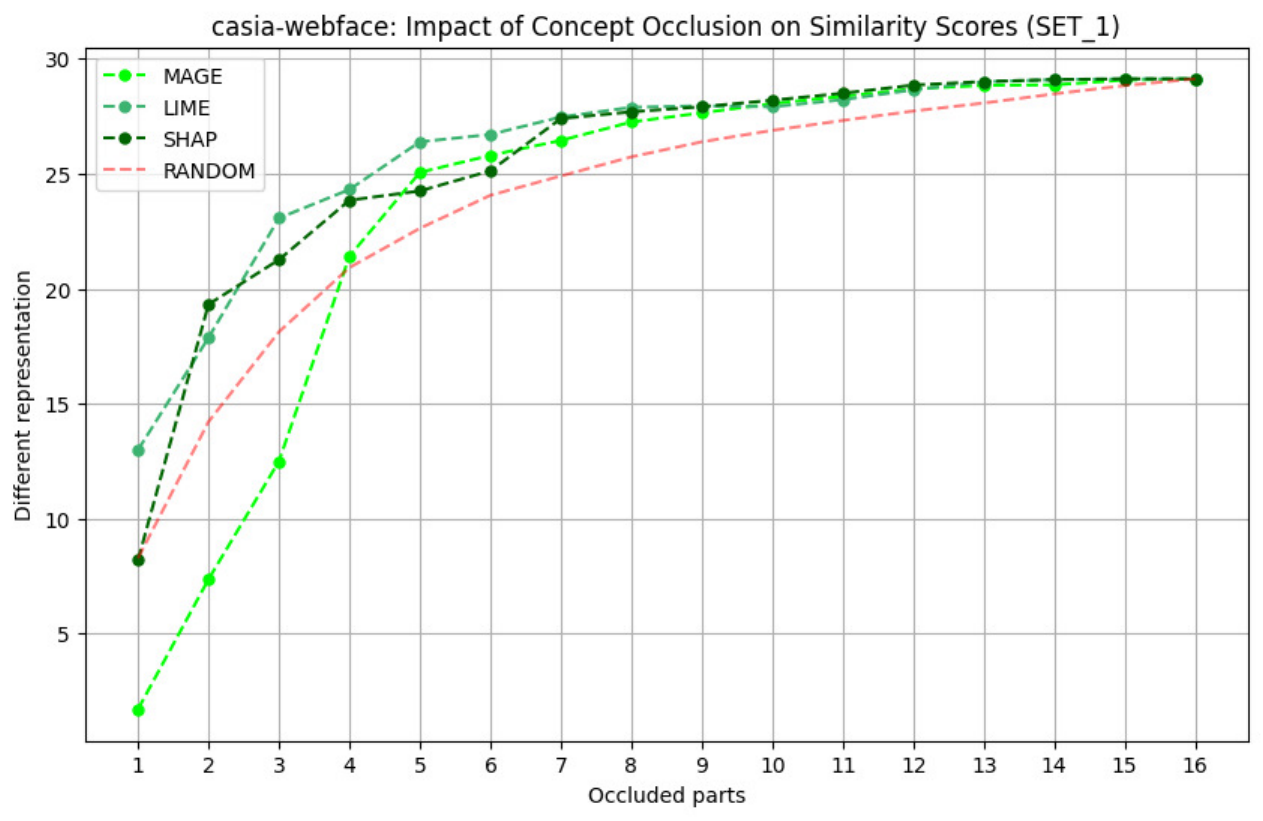} \label{fig:image5_sim}}
     \\
     \subfloat[]{\includegraphics[width=0.45\linewidth]{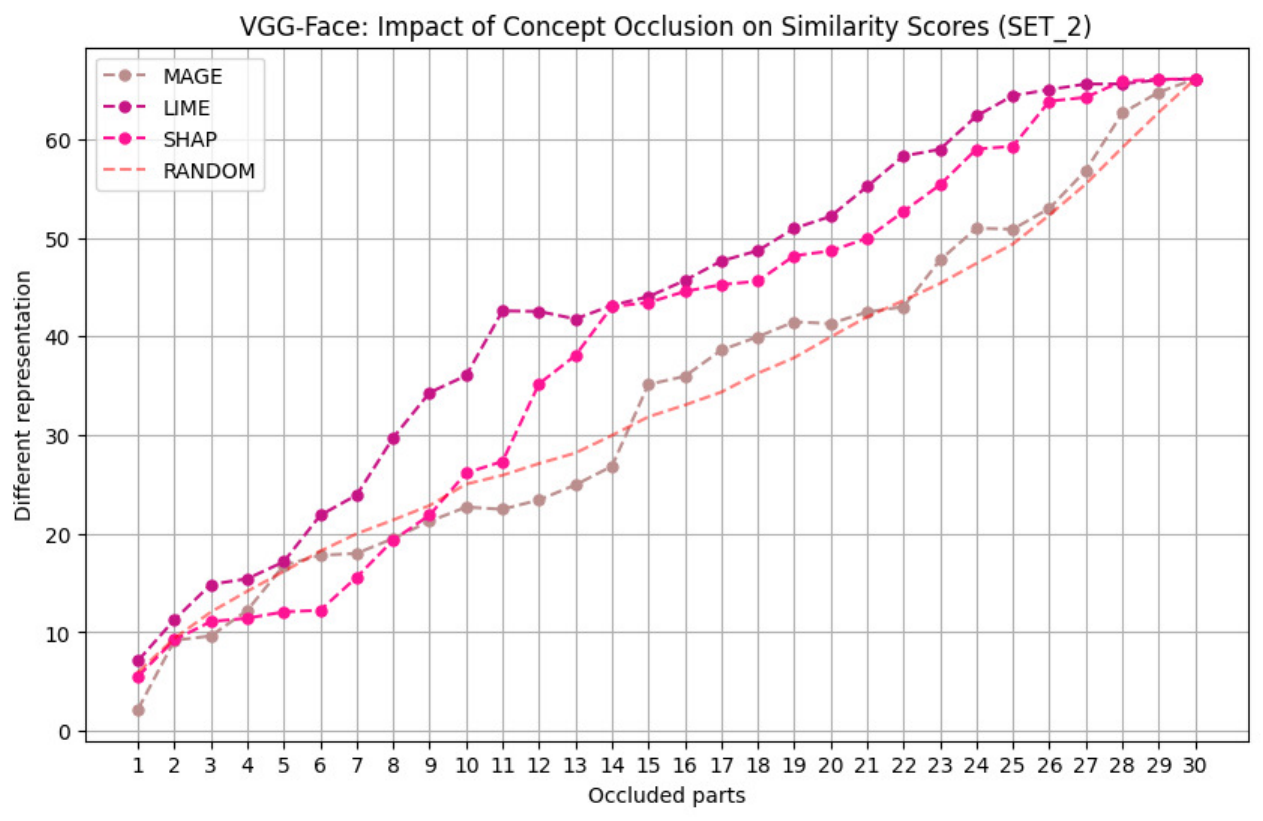} \label{fig:image3_sim}}~
     \subfloat[]{\includegraphics[width=0.45\linewidth]{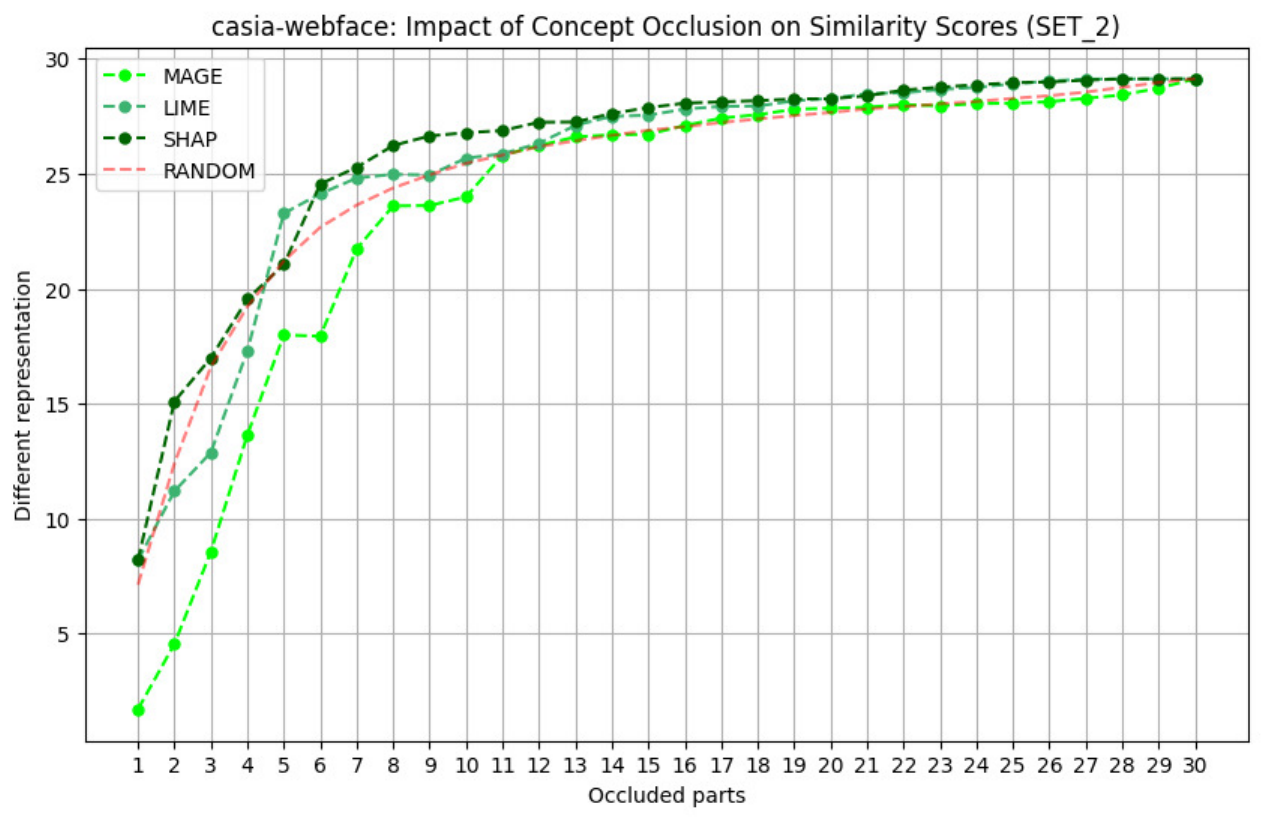} \label{fig:image6_sim}}
     \caption{Difference values of similarity score by successive occluding from the most to the least important concepts found for three different semantic sets of face regions, for VGGFace (a, c, and e) and Casia (b, d, and f) models. The x-axis presents the number of occluded parts, and the y-axis presents the distance between the similarity score of the original pair of face images and, the pair with one occluded face image. We show the comparison of three xAI techniques to find concepts (MAGE, LIME and SHAP) based on three semantic sets. We compare the results with a randomly based occlusion of the concepts for the three semantic sets. The random approach represents an average behavior for choosing concepts, and superior values represent good performances.}
     \label{fig:occ_similarity_vgg}
 \end{figure}

For the face verification task comparison, we compare the difference between the similarity scores of 350 pairs of images before and after occlusion using the Euclidean distance. We present the results for CasiaWebFace and VGGFace2 trained models in Figure~\ref{fig:occ_similarity_vgg}.

We noticed four main things: all the methods can better perform than the random behavior at some point; the best method for this type of occlusion-based experiment is LIME for all the models and sets; given the decomposition nature (dividing the network into clusters of concepts) of the MAGE technique, it can reach good results only by combining the occlusion of more concepts (representing different parts of the network) which means at the beginning it performs poorly, but it can outperform SHAP after a few concepts for some models; SHAP has an average behavior as it is a local explanation and, as the tested images are different from the used for the concepts' extraction, it will only behave correctly if the network is well-generalized.

\subsection{Local Explanation and Sensitivity Analysis}
\begin{figure*}[!ht]
    \centering
    \includegraphics[width=\linewidth]{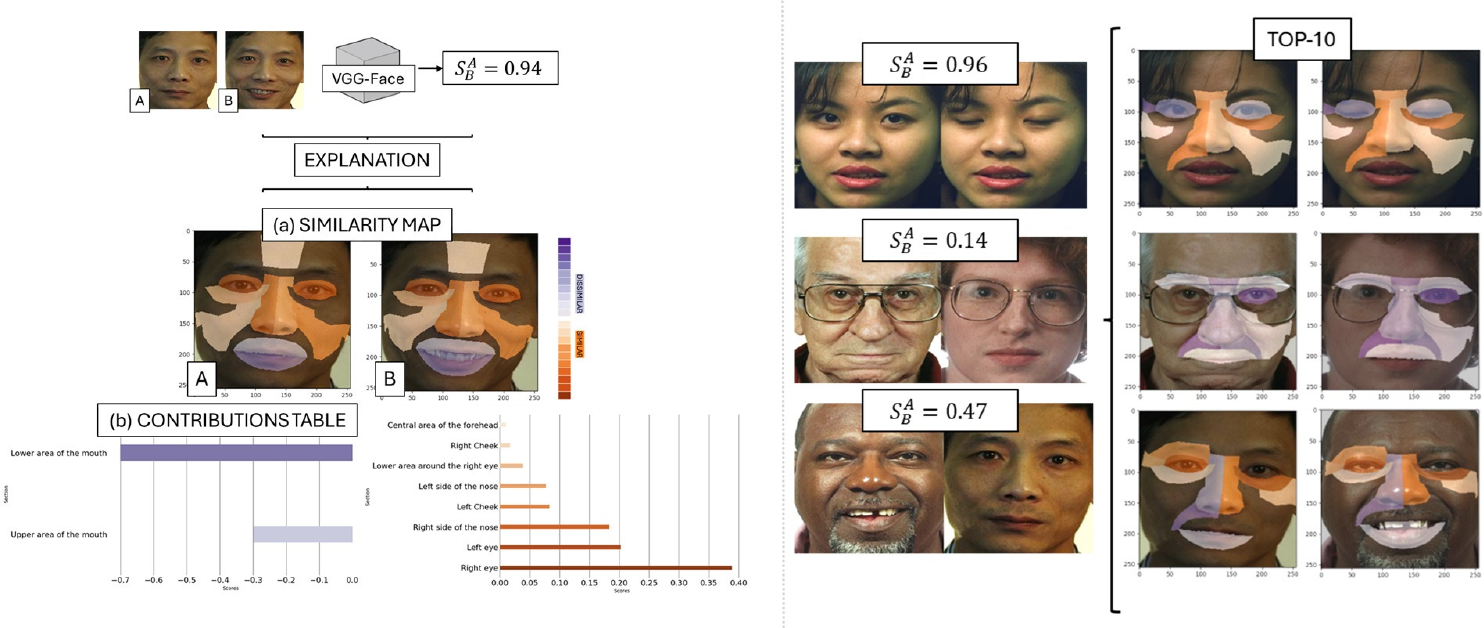}
    \caption{(Left) Example of local explanation for VGG-Face model output ($S_{AB} = 0.94$) showing similarity maps and a contribution table for semantic features from SET\_2. (Right) Similarity maps for various cases, highlighting the top 10 features with the greatest local impact, balanced by their global importance.}
    \label{fig:ex}
\end{figure*}
In this section, we present an example of the output for local explanations generated by our framework. Specifically, Figure \ref{fig:ex} (left side) illustrates an explanation for images A and B for the VGG-Face model output ($S_{AB} = 0.94$). The figure includes two similarity maps showing semantically similar (orange range) and dissimilar (purple range) areas, alongside a contribution table that highlights these semantic areas, displaying the magnitude of their similarity or dissimilarity. This example utilizes the semantic features from SET\_2, the most detailed set among those hypothesized. Due to space constraints, we only display the similarity map and the contribution table, which are the results of the previously presented single removal algorithm. We will address the textual explanation in the user feedback section.

On the right side of the figure, similarity maps for various cases are presented, ordered by true positives (such as two images of the same individual) and impostor pairs (such as two different individuals compared). These similarity maps provide detailed explanations. In the first case, the images are nearly identical except for the subject closing their eyes in one image. The similarity map identifies the eyes as dissimilar areas, while the rest of the face remains similar. In the second case, involving two different individuals, all features are marked as dissimilar. The third case is more intriguing: although the subjects differ in features, such as the left side of the nose and the lower lip, the model perceives the upper facial features as similar, explaining the similarity score of 0.47 despite the significant differences.

In these examples, we highlight the top 10 most important features. Unlike in our previous work, the top 10 features here are not selected based solely on the model’s global behavior but are those with the greatest local impact (in absolute value) after being balanced with their global importance. This approach visualizes features with local importance while considering the model’s overall behavior.

\begin{figure}[!ht]
    \centering
    \includegraphics[width=0.8\linewidth]{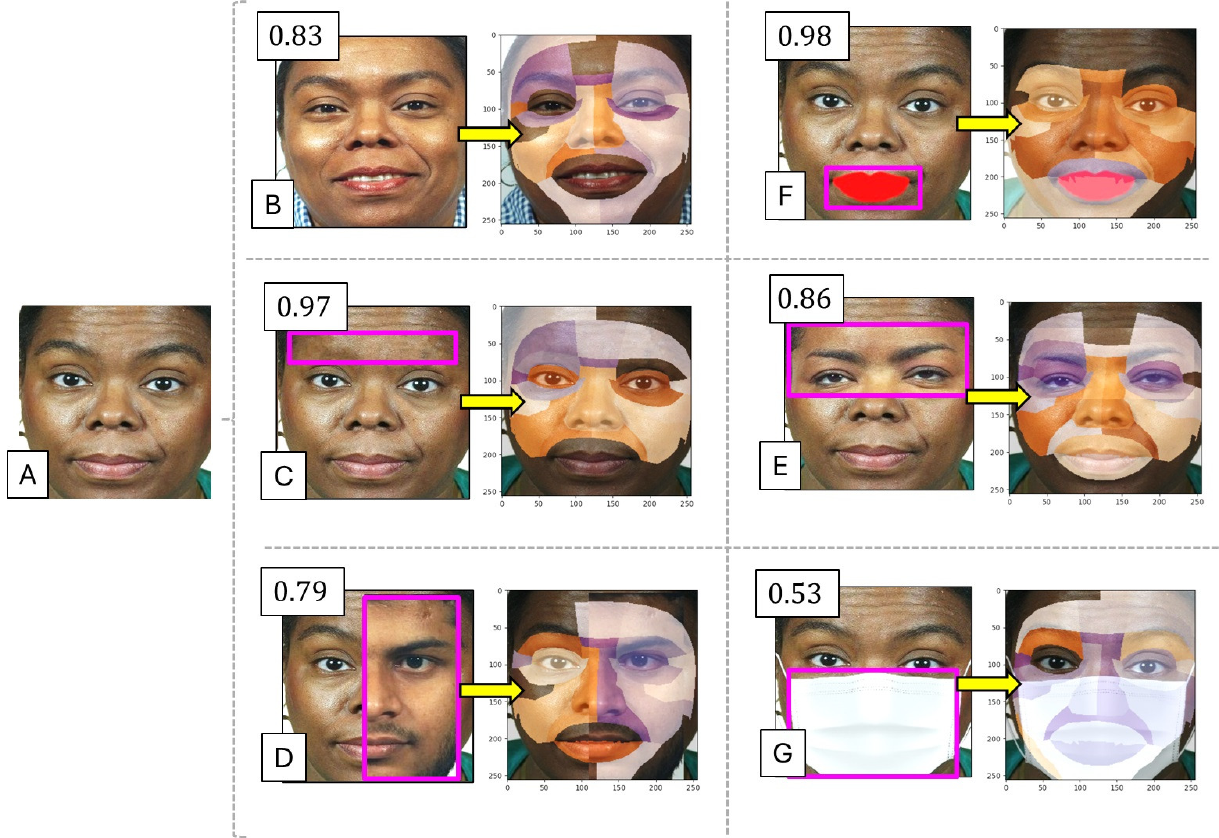}
     \caption{Example of local explanations for the VGG-Face model with their respective similarity scores. (A) Original image. (B) True positive image. (C) Image with eyebrows removed. (D-E) Images with parts from another individual's face (D: eyes and forehead, E: half face). (F) Image with a fake mouth added. (G) Image with a mask occlusion. The similarity maps indicate semantically similar (orange) and dissimilar (purple) areas, with the top 20 most important features highlighted for cases B-F, and the top 25 features for case G.}
    \label{fig:sens}
\end{figure}

As in previous work~\cite{Doh2024Beware}, we conducted a sensitivity analysis using a “Cut-and-Paste” experiment to test the robustness of this method  \ref{fig:sens}. Previously, we replaced specific facial regions in an image with corresponding regions from another image to detect high similarity in the similarity maps, as in the cases of copies AD (half-face) and AE (eyes), where parts of another person's face were pasted onto the original sample. In this study, we opted for a similar experiment but with variations in the type of modifications applied to the image. Starting with a true match (A-B), we then applied modifications to the original image. For instance, in image C, the eyebrows are removed; in image F, a fake mouth is pasted onto the image; and in image G, a facial mask occlusion is applied. For all cases, the top 10 features are shown, while for the facial mask occlusion case, the top 25 features are displayed to demonstrate if the eyes are recognized as similar.

As observed, the similarity maps are sensitive to changes not only in terms of features but also in detecting similarities and dissimilarities. When discrepancies arise, such as recognizing dissimilar areas that actually belong to image A, this can be attributed to the holistic perception of faces by network models. Altering a specific patch may lead to a change in the perception of the entire face, not just the modified area. This explanation aligns with the study by Jacob et al.\cite{Jacob2021}, which demonstrated through the Thatcher effect \cite{Margaret1980}—a phenomenon where local changes to a face, such as inverting the eyes or mouth, are challenging to detect when the face is presented upside down but become strikingly apparent when the face is right side up—  that models trained on various face datasets internalize a holistic perception of faces.

\subsection{User Feedback Evaluation}
In this section, we delve into the findings from our user survey aimed at evaluating the interpretability and effectiveness of our framework. The survey was designed to gather feedback on several key aspects, including the types of semantics used, the visualization methods, and the inclusion of textual explanations. Our objective was to understand how users perceive the different components of our framework and to identify areas for improvement based on their feedback. This analysis provides insights into user preferences and experiences across varying technical backgrounds, informing the refinement of our approach to enhance interpretability and satisfaction.

\subsubsection{Participant Demographics}

\begin{figure}[!ht]
    \centering
    \includegraphics[width=0.7\linewidth]{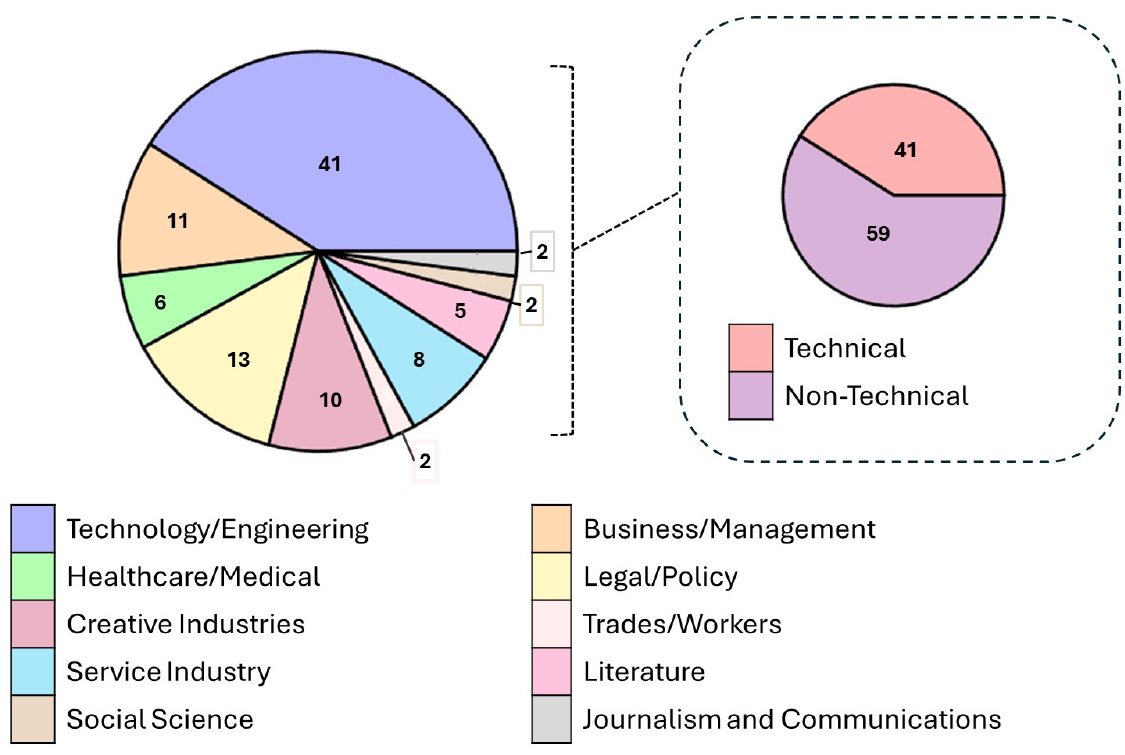}
     \caption{Distribution of participants' professional backgrounds categorized into technical (41\%) and non-technical (59\%) roles. The technical group includes fields such as technology, engineering, and related disciplines, while the non-technical group encompasses diverse areas, including healthcare, creative industries, business, and more}
    \label{fig:professional_background}
\end{figure}

After evaluating the framework, we decided to assess the interpretability of the explanations through a Google Form survey, as previously introduced in Section \ref{sec:userevaluation}. Our survey gathered feedback from 61 individuals over the course of one week. The distribution of the professional backgrounds of the participants is illustrated in the pie charts in Figure  \ref{fig:professional_background}).

The objective was to gather feedback from a diverse audience, encompassing various professional backgrounds. Participants identified their professional background, which we categorized into two main groups: technical and non-technical. The technical group comprised 41\% of the respondents, including those with a background in technology or engineering. The non-technical group, representing 59\%, included all other professional backgrounds, resulting in a balanced distribution.

Additionally, we collected further demographic data for a more comprehensive understanding of our participants. The majority of respondents were young adults, with 72\% aged between 25 and 34 years. Smaller age groups included 10\% aged 18-24, 7\% aged 35-44, 5\% aged 55-64, and 2\% for both 45-54, 65 and above, and under 18 years. This concentration in the 25-34 age range could reflect a potential sampling bias related to the authors' ages.

Regarding educational attainment, the respondents were predominantly highly educated: 55\% held a Master's degree, 16\% had completed high school or an equivalent, 14\% had a Bachelor's degree, and 5\% had primary education.

The survey also revealed a broad range of familiarity with AI among participants: 36\% had a basic understanding of AI, 20\% considered themselves very knowledgeable, 17\% had moderate knowledge, 15\% had no knowledge of AI, and 12\% were experts in the field.

Responses to understanding XAI varied significantly. While 32\% of participants had a basic understanding of XAI, 28\% had never heard of it. Another 18\% had heard of it but did not know what it meant, 15\% understood it well enough to explain it to others, and 7\% were highly knowledgeable with practical experience.

To ensure a comprehensive analysis, we examined the survey responses by categorizing participants into technical and non-technical groups based on their professional backgrounds. This approach provided a clear perspective on understanding the explanations and effectively balanced the groups. While it was possible to analyze each demographic aspect individually, doing so would have been overly complex and lengthy. Therefore, focusing on the distinction between technical and non-technical backgrounds was a practical and meaningful basis for our analysis.

\subsubsection{Framework and explanation evaluation}
To comprehensively evaluate the proposed framework, we asked our users to assess various aspects, including the type of semantics, the visualization of the explanations, and the inclusion of textual transcription.\\
\textbf{Preference in Semantic:}
Collecting detailed semantics and definitions from all users was computationally time-consuming. Therefore, our experiments were based on three hypothetical semantics, which were later evaluated by the users. From the user feedback, four properties were frequently mentioned:
\begin{itemize}
\item Completeness: The extent to which a semantic set provides a detailed and comprehensive analysis, covering all relevant areas and features of the face.
\item Clarity: The extent to which a semantic set is easy to understand and interpret for users.
\item Precision: The extent to which a semantic set segments facial features with a high degree of specificity and accuracy.
\item Simplicity: The extent to which a semantic set is straightforward and easy to use, avoiding unnecessary complexity.
\end{itemize}

Based on these properties, we analyzed the users' responses. Semantic set 2 was the most preferred (50\%), while sets 1 and 0 had nearly equal preferences of 25\% and 26\%, respectively.

\begin{table}\caption{ Overview of semantic evaluations based on four properties: Completeness, Clarity, Precision, and Simplicity. Each semantic set (SET\_0, SET\_1, SET\_2) is evaluated by users, with percentages indicating the preference for each property. SET\_0 is primarily chosen for its simplicity, SET\_1 for its balanced approach, and SET\_2 for its completeness and precision.}
\centering
\pgfplotstabletypeset[color cells]{
x,Completeness,Clarity,Precision,Simplicity
SET\_0,22.22,33.33,22.22,77.78
SET\_1,28.57,28.57,14.19,28.57
SET\_2,62.50,18.75,68.75,6.25
}
\label{tab:semanticpreference}
\end{table}

As shown in Table \ref{tab:semanticpreference}, SET\_0 was primarily chosen for its simplicity. One user noted, ``I prefer SET\_0 as it is simpler and more flexible.'' SET\_2 was appreciated for its completeness and precision. A user commented, ``SET\_2 because the granularity is higher so I can have higher accuracy in space.'' SET\_1 was seen as the most balanced. One user explained, ``I chose SET\_1 because it feels closer to me when I evaluate similarities in facial features."

We observed differences in preferences based on the participants' backgrounds:
\begin{itemize}
    \item Technical Background: Preferences were distributed: 35\% for SET\_2, 30\% for SET\_1, and 26\% for SET\_0. Technical users emphasized precision and completeness. For example, a technical user stated, ``I chose SET\_0 because the separation of the human face into its basic parts is sufficient to study the similarity."

    \item Non-Technical Background: Non-technical participants showed a stronger preference for SET\_2 (61\%), followed by SET\_1 (22\%) and SET\_0 (11\%). These users valued simplicity and clarity. A non-technical participant commented, "SET 2: It contains more details, making it seem more accurate."
\end{itemize}

These observations indicate that while SET\_2 is generally preferred, technical users appreciate the detailed aspects of the semantic sets, whereas non-technical users favor simplicity and ease of understanding.\\
\textbf{Clarity of the Visualization:}
Most participants rated the explanations as fairly or extremely clear. Among those with a technical background, approximately 76\% found the explanations clear, 19\% were neutral, and 5\% found them somewhat unclear. Similarly, 74\% of non-technical participants found the explanations clear, 18\% were neutral, and 9\% found them somewhat unclear. None of the participants, regardless of their background, rated the explanations as not clear at all.

Technical participants generally showed higher levels of satisfaction. Over three-quarters were fairly satisfied, and about 10\% were very satisfied. Around 14\% were neutral, and none were dissatisfied. In contrast, non-technical participants exhibited a broader range of opinions. While 35\% were fairly satisfied and 24\% very satisfied, about a third were neutral, and approximately 9\% were somewhat unsatisfied. None of the non-technical participants were completely dissatisfied.

Regarding the clarity of similarity values presented in the table, all technical participants found the values clear, with 86\% rating them as fairly or extremely clear and 14\% neutral. This unanimous agreement highlights the table's effectiveness for this group. Among non-technical participants, most found the similarity values clear, although about 6\% did not. One participant noted, "I am not very familiar with tables. I understand, for example, that the lower area around the right eye is very dissimilar and the right side of the nose is similar, but I cannot understand how you got the 64\% value." This indicates that while the majority found the values clear, a small portion struggled with understanding them.

The utility of the table for understanding the analysis was also evaluated. Among those with a technical background, 95\% found the table useful, with only 5\% indifferent and none finding it not useful. For non-technical participants, 74\% found the table useful, 18\% were indifferent, and 9\% found it not useful. This indicates a slightly lower level of perceived utility among non-technical users than their technical counterparts.

Technical participants who provided feedback on unclear aspects of the similarity table often suggested detailed improvements for better visualization. They demonstrated a clear understanding of the values but recommended aesthetic enhancements. For instance, one participant mentioned, ``I find the table clear, but it would be more visually pleasant with some modifications." Another suggested, ``It would be interesting to visualize similarity values only by clicking on the area of interest without having the table displayed directly."

Non-technical participants who found the values unclear cited difficulties in understanding the percentages and issues with optimization for mobile devices. One participant stated, ``I am not very familiar with tables. I understand, for example, that the lower area around the right eye is very dissimilar and the right side of the nose is similar, but I cannot understand how you got the 64\% value." Another noted, ``It is an image not optimized for smartphones. Considering that most people do this on mobile, every image should be optimized for mobile." This feedback underscores the need for improved clarity and accessibility in the table's presentation.

Overall, most participants in both groups found the explanations clear, with technical participants slightly preferring higher clarity ratings. Technical participants also demonstrated greater overall satisfaction with the method than non-technical participants, who exhibited more varied opinions. Both groups found the table useful for understanding the analysis, with a higher majority among technical participants. While all technical participants found the similarity values clear, a small percentage of non-technical participants did not. Some non-technical participants expressed difficulties in understanding specific values in the table.

Participants who found the explanations clear and were satisfied with the method also tended to find the similarity values in the table clear. This connection is particularly evident in the detailed feedback from technical participants. Non-technical participants who did not find the similarity values clear often cited general comprehension issues and optimization problems, indicating a link between overall explanation clarity and the perception of the table. These results highlight the importance of tailoring explanations and visualizations to meet the needs of users with varying technical expertise.

\textbf{Textual Transcription:} As mentioned in Section \ref{sec:textrans}, we used automatic textual transcription of similarity values provided by Codellama instruct-7B, Zephyr-7B-beta, and Beagle14-7B. We show the preferences in Table~\ref{tab:semanticpreference2}. Each model demonstrated distinct stylistic approaches to explain the face verification model's decisions, informed by a detailed prompt. Due to space constraints, the full-text outputs of each model are reported in the \ref{sec:appendix_A}.

Codellama instruct-7B offers a detailed and practical explanation, focusing on specific facial features to illustrate similarities and dissimilarities between two images. For example, it states, ``Specifically, it (\textit{the model}) is seeing differences in the right eye, left eye, upper area of the mouth, central area of the forehead, and right cheek.” It also uses relatable examples: “For instance, if you have a photo of your grandmother from one year ago and another photo of your grandmother now, the model will see her nose as dissimilar while seeing some features (like the eyes) as similar.” This practical style aids comprehension, especially for users with non-technical backgrounds.

Zephyr-7B-beta provides a balanced and concise explanation detailing how different facial regions contribute to the similarity score. It explains, ``Areas with dissimilarities tend to lower the overall cosine similarity score, while areas with similarities tend to increase it.” This model avoids excessive technical jargon, making the explanation accessible and informative for both technical and non-technical audiences.

Beagle14-7B employs a structured and detailed style, utilizing visual aids like color-coded maps to indicate levels of similarity and dissimilarity. It describes, ``The color map will display shades of orange where those similarities are found (i.e., higher scores), while shades of purple indicate differences or dissimilarities (lower scores).” This model offers comprehensive transcriptions without numerical values for specific areas, such as, ``The 'Left side of the nose' and 'Lower area around the right eye,' for example, have lower similarity values indicating that they were seen as less alike in both images. Overall, the final cosine similarity score combines the contributions made by each facial area.” The organized presentation and visual aids enhance clarity and understanding.

User preferences varied based on the clarity, conciseness, and structure of each transcription style. Codellama instruct-7B is preferred for its clear presentation and practical examples, making technical aspects more accessible. This style is particularly beneficial for users with non-technical backgrounds who appreciate the simplicity and accessibility of the explanations.

Zephyr-7B-beta is appreciated for its balanced and concise explanations. Users value its ability to provide essential information briefly, making it suitable for quick comprehension. This model appeals to both technical and non-technical audiences.

Beagle14-7B is the most preferred transcription style, praised for its well-organized text structure and use of visual aids. It is considered more descriptive and inclusive of valuable details, enhancing understanding. Users with technical backgrounds particularly prefer this model for its clear and organized explanation of the evaluation method, while non-technical users find the color-coded maps very helpful for clarity.

Overall, Beagle14-7B's structural clarity and informational detail make it the most preferred choice among users, aligning with the needs of both technical and non-technical backgrounds. This highlights the importance of clear, concise, and well-structured explanations in understanding AI mechanisms.
\begin{table}\caption{Comparison of preferences for model transcription styles between participants with technical (T) and non-technical (NT) backgrounds. The table illustrates the percentage of users who preferred each transcription style: Beagle, Zephyr, and Codellama. Beagle is the most preferred style among both technical (42.86\%) and non-technical (44.12\%)}
\centering
\pgfplotstabletypeset[color cells]{
x,Beagle,Zephy, Codellama
T,42.86,23.81,32.35
NT,44.12,23.53,31.30

}
\label{tab:semanticpreference2}
\end{table}

\subsection{Proposed method vs. Traditional one}

\begin{figure}[ht]
    \centering
    \includegraphics[width=0.8\linewidth]{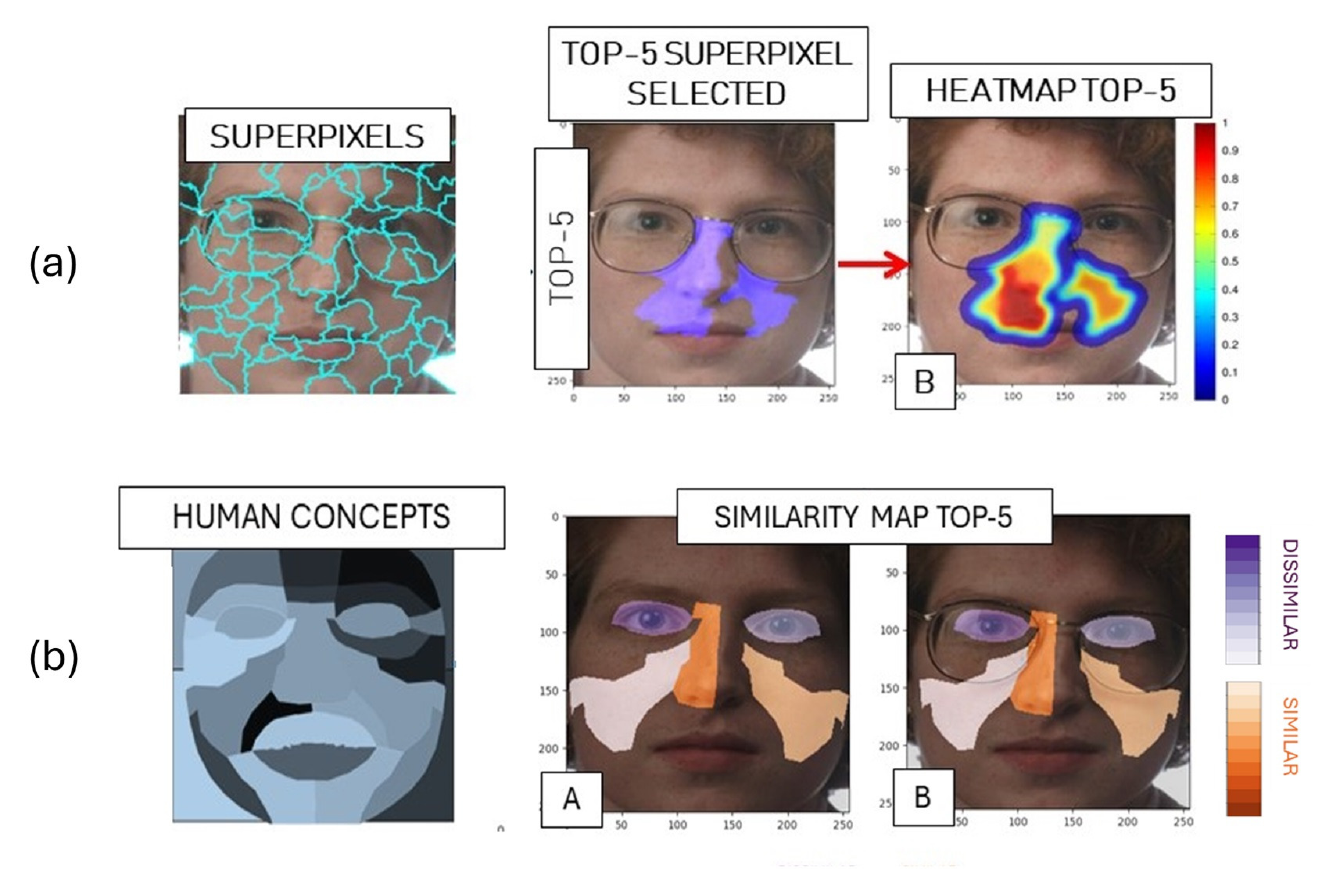}
    \caption{(a) Traditional method represented by LIME using superpixels and standard heatmap's colours to represent the explanation (b) Proposed method using human concepts with Similar and dissimilar colours to represent the explanation. Both methods are represented with the top 5 most important concepts.}
    \label{fig:enter-label}
\end{figure}

An additional question asked to users was to express a preference between the semantic method proposed in the framework and a traditional method. For the traditional method, we opted to use LIME in its most common visual form, namely superpixels and heatmaps (with red indicating more important areas and blue indicating less important ones), as show in Figure \ref{fig:enter-label}.a, providing a brief explanation to help users understand the use of colours and the division into superpixels.

Among the analysed group of users, there was a clear preference for the semantic approach (78\%) compared to the traditional approach (12\%). The main reasons for this preference can be summarised into four main categories (technical and not technical distribution shown in Tab.\ref{tab:traditionvssemantic})
Firstly, users appreciated the level of detail and clarity provided by the semantic approach. They noted that this method ``provides more details'' and ``gives a clearer picture of the metrics''. Furthermore, it was observed that the semantic method is ``better for understanding specific features''.

Secondly, many users highlighted that the reasoning required by the semantic method is closer to how people normally compare two faces. One user stated that ``it's closer to the way I reason when comparing two faces'', while another found that ``the semantic approach seems clearer since the different parts of the face are easier to identify and compare''.

Thirdly, the semantic method was perceived as more logical and understandable. Users mentioned that ``the semantic approach is linked to how humans attribute meaning'' and that it ``explains similar areas better''.

Finally, the semantic approach offered a clearer visual mapping. As one user noted, ``the location of importance is clearer'' and ``the semantic approach seems clearer as it breaks down the interest/analyzed areas and scores them''.

For those users who preferred the traditional method, the main reasons revolved around its simplicity and familiarity. They indicated that it is ``easier to understand quickly'' and that it is ``traditional, easier to understand.'' Some users felt that ``the traditional method is clearer than the semantic one.''

Additionally, some users perceived the traditional method as providing a more accurate representation. One user commented: ``I chose the traditional method rather than the semantic approach. It's more relevant. I don't care about trying to `translate' an AI's methods into `human-friendly terms' if it's less accurate as to what it’s actually doing''.

Finally, the traditional method was appreciated for its clear visual impact. One user stated that ``the traditional approach: colored areas have a very clear visual and comprehension impact'', while another simply mentioned ``traditional approach with heatmaps.''

In conclusion, the preference for the semantic method seems to derive from its ability to provide richer details and greater clarity and reflect a reasoning process closer to human thought. On the other hand, the familiarity and simplicity of the traditional method and the perception of a more immediate and accurate visual representation led some users to prefer it. The choice between the two methods thus appears to depend on the balance between seeking detailed clarity versus visual simplicity and familiarity.The increasing complexity of machine learning models in computer vision, particularly in face verification, necessitates the devel-
opment of explainable artificial intelligence (XAI) to enhance interpretability and transparency. This study extends previous work
by integrating semantic concepts derived from human cognitive processes into XAI frameworks to bridge the comprehension gap
between model outputs and human understanding. We propose a novel approach combining global and local explanations, using
semantic features defined by user-selected facial landmarks to generate similarity maps and textual explanations via large language
models (LLMs). The methodology was validated through quantitative experiments and user feedback, demonstrating improved
interpretability. Results indicate that our semantic-based approach, particularly the most detailed set, offers a more nuanced un-
derstanding of model decisions than traditional methods. User studies highlight a preference for our semantic explanations over
traditional pixel-based heatmaps, emphasizing the benefits of human-centric interpretability in AI. This work contributes to the on-
going efforts to create XAI frameworks that align model cognition with human cognitive processes, fostering trust and acceptance
in critical applications.wever more than 90\% of technical people and more than 70\% of non technical prefer the semantic approach. 

\begin{table}\caption{Comparison of Technical (T) and Non-Technical (TN) Background Preferences for XAI method visualisation}
\centering
\pgfplotstabletypeset[color cells]{
x,Traditional,Semantic
T,2.40,90.48
NT,29.41,70.59

}
\label{tab:traditionvssemantic}
\end{table}
\section{Conclusion}
\label{ch:conclusion}
In this study, we proposed a novel framework for enhancing explainability in AI models used for face verification by integrating human-centric semantic approaches. Our approach extends traditional XAI methods by combining global and local explanations derived from user-defined semantic facial features. We validated the framework through both quantitative experiments and user feedback, intending to demonstrate its ability to improve interpretability. 
Key innovations include the introduction of several enhancements over our previous work: we expanded the framework by incorporating a more refined strategy for combining global and local explanations, adapting context-aware algorithms (such as LIME and MAGE) to extract global concepts from the models. Additionally, we quantitatively evaluated the importance of these concepts through occlusion experiments, a method not previously explored. We also integrated Large Language Models (LLMs) to automatically generate textual explanations, offering a more accessible interpretation of the model's outputs. Lastly, we conducted an extensive user study to assess the impact of our approach on user perception, marking a significant expansion from the initial conceptual framework presented earlier.

With the semantic extraction experiments, we noticed that global-aware methods such as LIME can present better behavior for finding important semantic face regions. Local explanation methods behave well when analyzing similar images, but may not be stable for a global analysis even after ranking aggregation.

The survey results showed that our semantic approach was positively received by participants, who recognized its logic and utility, especially in terms of visualizing explanations through semantic similarity maps. Users appreciated having a more intuitive and detailed explanation than traditional methods, such as superpixel-based heatmaps. However, some participants highlighted the need for further refinement in the level of detail in the explanations, particularly to make semantic concepts clearer and improve the accessibility of the information.

Compared with the traditional method, the majority of participants preferred our approach, finding it clearer and more aligned with their way of reasoning about facial similarities. Despite this success, we must acknowledge that our survey sample was limited to 61 people, and there may be a ``bubble'' effect due to the similarity in age and background between the authors and some participants. Nevertheless, the feedback collected provides valuable insights for further improving the framework and optimizing the user experience.

Future work will focus on refining the semantic sets to further enhance clarity and usability, particularly for non-technical audiences. Additionally, we plan to explore the integration of more advanced LLMs to improve the quality and relevance of the textual explanations. Expanding the framework to include more diverse datasets and testing its applicability across different AI tasks will also be crucial steps in advancing the utility and acceptance of explainable AI in critical applications.

\bibliographystyle{elsarticle-num}
\bibliography{refs}
\clearpage
\appendix

\section{Additional Material}
\label{ch:appendix}
\subsection{Table of Facial Areas and Input Values Used for Model Testing}
\label{sec:appendix_A}
\begin{figure}[h!]
    \centering
    \includegraphics[width=0.7\linewidth]{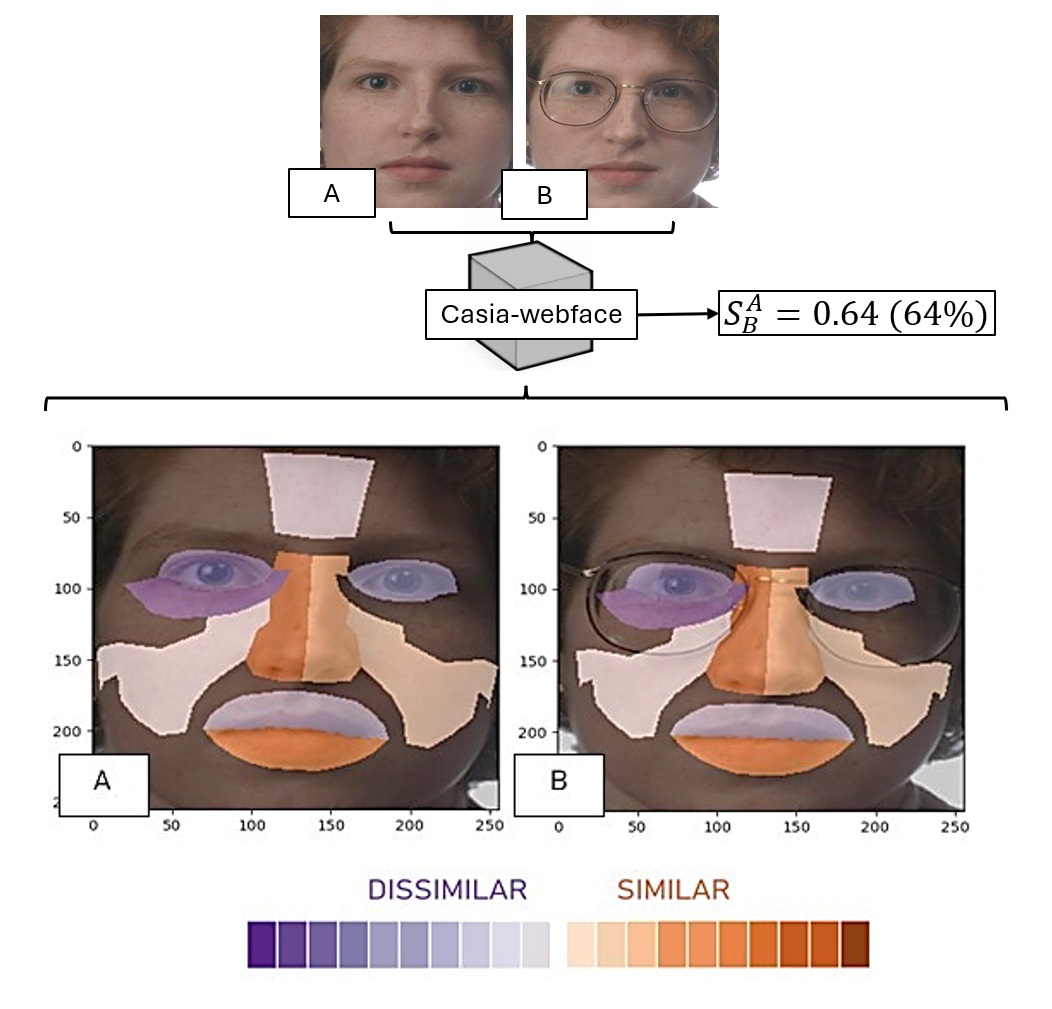}
    \caption{Visual explanation associated with the contribution table given as input to the LLMs}
    \label{fig:exprop}
\end{figure}

\begin{table}[h!]
\centering
\resizebox{0.5\textwidth}{!}{%
\begin{tabular}{|l|l|l|l|}
 \hline
 \multicolumn{2}{|c|}{\textcolor{violet}{Negative}} & \multicolumn{2}{|c|}{\textcolor{orange}{Positive}} \\ \hline
'Lower area around the right eye' & -0.0041 & 'Left Cheek'               & 0.0001  \\ \hline
'Right eye'                       & -0.0039 & 'Left side of the nose'    & 0.0003  \\ \hline
'Left eye'                        & -0.0024 & 'Lower area of the mouth'  & 0.0003  \\ \hline
'Upper area of the mouth'         & -0.0005 & 'Right side of the nose'   & 0.0010  \\ \hline
'Central area of the forehead'    & -0.0002 &                            &         \\ \hline
'Right Cheek'                     & -0.0001 &                            &         \\ \hline
\end{tabular}
}
\caption{Example of the table with facial areas and values given as input to the LLMs.}
\label{tab:inputdata} 
\end{table}

\subsection{Generated Outputs from LLMs showed in the survey}

The following sections report the textual explanations generated by three different large language models (CodeLlama instruct-7B, Zephyr-7B-beta, and Beagle14-7B) in response to the prompt described in section \ref{sec:textrans}, specifically for the case shown in Figure\ref{fig:exprop} and Table \ref{tab:inputdata}. Each model received the same prompt and produced slightly different interpretations.
\subsubsection{Codellma instruct-7B}
\label{sec: Appendix_A.2.1}
\texttt{ ``Based on the model’s output (cosine similarity) and its capabilities, we can explain that the face verification system sees two images as very similar in terms of features related to human-understandable concepts. Specifically, it is seeing differences in the right eye, left eye, upper area of the mouth, central area of forehead, and right cheek. The model sees these areas in the two images as similarities (higher similarity scores). On the other hand, it sees areas related to the nose, lips, and left side of the nose as differences or dissimilarities (lower similarity scores), which contribute to the overall [cosine\_similarity\_percentage] similarity score. The fact that these specific areas are very different does not mean that they are entirely irrelevant in determining the overall similarity between the two images. For instance, if you have a photo of your grandmother from one year ago and another photo of your grandmother now, the model will see her nose as dissimilar, while seeing some features (like the eyes) as similar. In conclusion, the face verification system sees these areas in the two images as very similar or dissimilar, which leads to a high overall similarity score of [cosine\_similarity\_percentage]’’ }
\\
\subsubsection{Zephyr-7b-beta:}
\texttt{ ``Based on the provided concepts and their associated scores, it can be observed that while some areas such as the left and right sides of the nose, lower area of the mouth, and central area of the forehead have similarities between the two images, other areas like the lower area around the right eye, right eye, left eye, upper area of the mouth, and right cheek show dissimilarities. The overall cosine similarity score of [cosine\_similarity\_percentage]suggests that while there are some differences between the two images, particularly in certain facial features, there are still notable similarities as well. These areas contribute differently to the final score based on their specific importance in distinguishing between faces. Areas with dissimilarities tend to lower the overall cosine similarity score, while areas with similarities tend to increase it. However, the exact impact of each semantic area on the final score is not explicitly stated and would require further analysis.’’ } 
\\
\subsubsection{Beagle14-7B:}
\texttt{ ``The [cosine\_similarity\_percentage] between two images indicates a moderate level of similarity, where certain facial features are seen as more or less alike in the respective images. The color map will display shades of orange where those similarities are found (i.e., higher scores), while shades of purple indicate differences or dissimilarities (lower scores).\\
The 'Left side of the nose' and 'Lower area around the right eye,' for example, have lower similarity values indicating that they were seen as less alike in both images. On the other hand, areas like the 'Right cheek,' 'Central area of the forehead,' or the 'Left cheek' showed slightly higher similarity scores, which means these features were more similar in both images.\\
Overall, the final cosine similarity score is a combination of the contributions made by each facial area. A difference or similarity in a specific area could either decrease (dissimilar) or increase (similar) the overall result, depending on its individual value. However, this breakdown provides only a high-level understanding as numerical values are not provided for the semantic areas' similarity/dissimilarity scores.'' }
\subsection{Analysis of Key Instructions and Progressive Removal Tests}
In this section, we outline the rationale behind including specific instructions in the prompt provided to the large language models and detail the results of progressively removing these instructions. The instructions were carefully designed to ensure clear, consistent, and interpretable responses across all models tested. Below, we analyze how each instruction contributes to the overall quality of the outputs and discuss the outcomes when these instructions are omitted.

\subsubsection{Inclusion of ``(it’s a percentage from 0 to 100\%)" and ``($\geq$0), example: -0.5)")}
The inclusion of these specific instructions was crucial for improving the clarity, accessibility, and relevance of the outputs produced by the models. By specifying that the cosine similarity score is a percentage from 0 to 100\% and providing an example of positive and negative values, the models were better able to deliver explanations in terms of practical, easily understandable metrics rather than abstract mathematical concepts.

For example, consider the output of CodeLlama instruct-7B when these instructions were included: 
\\
\begin{justify}
    \texttt{``Based on the model’s output (cosine similarity) and its capabilities, we can explain that the face verification system sees two images as very similar in terms of features related to human-understandable concepts. Specifically, it is seeing differences in the right eye, left eye, upper area of the mouth, central area of forehead, and right cheek
[...].'' 
}(Codellama instruct-7B)
\end{justify}
In this case, the model effectively explains how certain areas contribute to the 64\% similarity score, linking the specific facial features to the overall result. However, when these instructions are removed, the model's explanation becomes more technical and potentially confusing, focusing on abstract concepts such as "distance between the left and right eye" without clear reference to the provided similarity data.
\begin{justify}
    \texttt{``The cosine similarity score between the two images is 64\% because certain visual features of the human face are present and consistent across the two images, while others are not. In particular, there are differences in the distance between the left and right eye...
[...].'' 
}(Codellma instruct-7B (without instruction))
\end{justify}
The reference to ``distance" becomes ambiguous because the model cannot ``see" the images and only has the tabulated similarity values. Mentioning specific distances is likely an assumption rather than a data-driven fact.
Moreover, the model may conflate cosine distance with physical distance between facial features, which it cannot measure.

Without clear instructions, the model risks creating ambiguous or misleading explanations that are not directly supported by the input data

\subsubsection{Inclusion of ``No long explanation"}
The ``No long explanation" instruction was included to ensure that the models produced concise, focused outputs that conveyed the necessary information without overwhelming the user with excessive detail. 
For instance, with the ``No long explanation" instruction in place, the models provided concise summaries that effectively communicated the main points as shown in \ref{sec: Appendix_A.2.1}.
Without this instruction, however, the models sometimes generated unnecessarily long responses that could confuse users or distract from the essential information:
\begin{justify}
    \texttt{The cosine similarity between the two images is 64\%. The system compares the semantic areas in both images and assigns a weightage to each area based on its importance. When an area is similar, it is given a positive score. When an area is dissimilar, it is given a negative score. The final score of these similarity or dissimilarity between two images is then calculated by taking the average of all the areas' scores.[...]. These areas have a negative similarity score, indicating they are dissimilar between the two images. This is because these areas contain distinct features such as a different shape or color that are not present in the other image. On the other hand, areas with positive similarity scores indicated that the features were similar between the two images. For example, the right eye and left cheek have a high similarity score indicating they share similar features [...]. The system assigns a weightage to each area based on its importance, with areas considered more important assigned higher weightages. Therefore, the final cosine similarity score is calculated by taking an average of all the areas' scores with their respective weights applied. The conclusion is that the image verification system has accurately identified the two images as similar even though they contain minor differences in certain semantic areas. The system can be used to verify the identity of a person or object in various scenarios such as security, surveillance, or facial recognition.
} (Codellama instruct-7B (without instruction))
\end{justify}

\subsubsection{Limitations of LLMs textual transcriptions}
It is important to note that the outputs generated by the models  sometimes contain statements that do not align with the input data (Tab.\ref{tab:inputdata}), specifically the similarity and dissimilarity values provided in the table. For instance, in both the original and modified outputs, Zephyr Beta suggests that areas like the 'central area of the forehead' and 'right cheek' show similarities, even though their respective values in the table are negative, indicating dissimilarity.
This inconsistency may arise because the model does not have visual access to the images and relies solely on the tabulated values. In cases where the model mentions specific distances or physical features, such as ``the distance between the left and right eye," it could be making assumptions rather than providing a fact-based interpretation of the data. Without clear and structured instructions, the model risks generating explanations that go beyond the available data or offer misleading interpretations.

This highlights the importance of properly structuring prompts and providing specific guidelines to ensure that the model generates explanations that accurately reflect the input data, especially in tasks involving numerical or technical information.
\end{document}